%% file: 00-main.tex
\documentclass[11pt,a4paper]{article}
\usepackage[hyperref]{acl2021}
\usepackage{times}
\usepackage{latexsym}

\usepackage{microtype}

\aclfinalcopy %
\def\aclpaperid{2509} %

\usepackage{bm}
\usepackage[caption=false]{subfig}
\usepackage{adjustbox}
\usepackage{booktabs}
\usepackage{pifont}
\usepackage{multirow}
\usepackage{graphicx}
\usepackage[linesnumbered, ruled, vlined]{algorithm2e}
\usepackage{algpseudocode}
\usepackage{amsfonts}
\usepackage{array,amsmath,amssymb}
\usepackage{colortbl}
\usepackage{xcolor}
\usepackage{subfig}

\usepackage{tabularx}

\def\figref#1{Figure~\ref{fig:#1}}
\def\figlabel#1{\label{fig:#1}\label{p:#1}}

\def\tabref#1{Table~\ref{tab:#1}}
\def\tabsref#1{Tables~\ref{tab:#1}}
\def\tabrefbare#1{\ref{tab:#1}}
\def\tablabel#1{\label{tab:#1}\label{p:#1}}

\def\secref#1{\S\ref{sec:#1}}
\def\seclabel#1{\label{sec:#1}}
\def\eqref#1{Eq.~\ref{eqn:#1}}

\newcounter{notecounter}
\newcommand{\enotesoff}{\long\gdef\enote##1##2{}}
\newcommand{\enoteson}{\long\gdef\enote##1##2{{
\stepcounter{notecounter}
{\large\bf \hspace{1cm}\arabic{notecounter} $<<<$ ##1: ##2 $>>>$\hspace{1cm}}}}}
\enoteson
\enotesoff                      
\def\zsx{ZS-XLT\xspace}
\def\fsx{FS-XLT\xspace}

\newcolumntype{Y}{>{\centering\arraybackslash}X}

\makeatletter
\def\thanks#1{\protected@xdef\@thanks{\@thanks
        \protect\footnotetext{#1}}}
\makeatother

\title{A Closer Look at Few-Shot Crosslingual Transfer: \\ The Choice of Shots Matters}

\author{
  Mengjie Zhao\textsuperscript{1}\textsuperscript{*}\thanks{\ *\ Equal contribution.}
  \ \ Yi Zhu\textsuperscript{2}\textsuperscript{*}
  \ \ Ehsan Shareghi\textsuperscript{3, 2}
  \ \ Ivan Vuli\'c\textsuperscript{2}\\
  {\bf Roi Reichart\textsuperscript{4}}
  \ \ {\bf Anna Korhonen\textsuperscript{2}}
  \ \ {\bf Hinrich Sch\"{u}tze\textsuperscript{1}}\\
  \textsuperscript{1}CIS, LMU Munich
  \ \ \textsuperscript{2}LTL, University of Cambridge\\
  \textsuperscript{3}Department of Data Science \& AI, Monash University\\
  \textsuperscript{4}Faculty of Industrial Engineering and Management, Technion, IIT\\
  {\tt mzhao@cis.lmu.de}, 
  {\tt \{yz568,iv250,alk23\}@cam.ac.uk},\\
  {\tt ehsan.shareghi@monash.edu},
  {\tt roiri@technion.ac.il}
}

\date{}

\begin{document}
\maketitle

\begin{abstract}
\input{01-abstract.tex}

\end{abstract}

\input{02-intro.tex}

\input{03-related_work.tex}

\input{04-method.tex}

\input{05-datasets_and_setup.tex}

\input{06-discussion.tex}
\input{07-conclusion.tex}

\section*{Acknowledgments}
This work was funded by the European Research Council:
ERC NonSequeToR (\#740516) and ERC LEXICAL (\#648909).
We thank the anonymous reviewers and Fei Mi for their helpful suggestions.

\bibliography{custom}
\bibliographystyle{acl_natbib}

\clearpage
\appendix   
\input{99-appendix.tex}

\end{document}

%% file: 01-abstract.tex
Few-shot crosslingual transfer
has been shown to outperform its
zero-shot counterpart with pretrained encoders like multilingual BERT.
Despite its growing popularity, little to no attention has been paid
to standardizing and analyzing the design of few-shot experiments.
In this work, we highlight a fundamental risk posed by
this shortcoming, illustrating
that the model exhibits a high degree
of sensitivity to the selection of few shots.
We conduct a large-scale experimental study on 40 sets of sampled few
shots for six diverse NLP tasks across up to 40 languages.
We provide an analysis of success and failure cases of few-shot
transfer, which highlights the role of lexical features.
Additionally, we show that a straightforward full model finetuning
approach is quite effective for few-shot transfer, outperforming
several state-of-the-art few-shot approaches.
As a step towards standardizing few-shot crosslingual
experimental designs, we make our sampled few shots publicly
available.\footnote{Code and resources are available at \url{https://github.com/fsxlt}}

%% file: 02-intro.tex
\section{Introduction}
Multilingual pretrained encoders like multilingual BERT
(mBERT; \citet{devlin-etal-2019-bert})
and XLM-R \citep{conneau-etal-2020-unsupervised}
are the top performers in crosslingual
tasks such as natural language inference \citep{conneau2018xnli},
document classification \citep{SCHWENK18.658,artetxe-schwenk-2019-massively},
and argument mining \citep{toledo-ronen-etal-2020-multilingual}.
They enable transfer learning through language-agnostic representations
in crosslingual setups \cite{hu2020xtreme}.

A widely explored transfer scenario is
\emph{zero-shot crosslingual transfer} \citep{
  pires-etal-2019-multilingual,conneau2019cross,artetxe-schwenk-2019-massively},
where a pretrained encoder is finetuned on abundant task data in the
source language (e.g., English) and then directly evaluated on
target-language test data, achieving surprisingly good performance
\cite{wu-dredze-2019-beto,hu2020xtreme}.
However,
there is evidence
that zero-shot performance reported in the literature has
large variance and is often not reproducible
\cite{keung2020evaluation,rios-mller-sennrich:2020:WMT}; the results
in languages distant from English fall far short of those similar to
English \citep{hu2020xtreme,liang2020xglue}.

\citet{lauscher2020zero} stress the importance of \emph{few-shot
  crosslingual transfer} instead, where the encoder is first finetuned
on a source language and then further finetuned with a small amount
(10--100) of examples (\textbf{few shots}) of the target language.
The few shots substantially improve model performance of the target
language with negligible annotation costs
\cite{garrette-baldridge-2013-learning,hedderich-etal-2020-transfer}.

In this work, however, we demonstrate that the gains from few-shot
transfer exhibit a \textit{high degree of sensitivity to the selection
  of few shots}.
For example, different choices for the few shots can yield a
performance variance of over 10\% accuracy in a standard document classification task. 
Motivated by this, we propose to fix the few shots for fair comparisons between different
crosslingual transfer methods, and provide a benchmark
resembling the standard ``$N$-way $K$-shot'' few-shot
learning configuration \citep{feifeilioneshot,Koch2015SiameseNN}.
We also evaluate and compare several state-of-the-art (SotA) few-shot
finetuning techniques, in order to understand their performance
and susceptibility to the variance related to few shots.

We also demonstrate that the effectiveness of few-shot
crosslingual transfer depends on the type of
downstream task. For
syntactic tasks such as named-entity recognition, the few shots can
improve results by up to $\approx$20 $F_1$ points.
For challenging
tasks like adversarial paraphrase identification, the few shots do not
help and even sometimes lead to worse performance than zero-shot
transfer.
To understand these phenomena, we conduct additional in-depth
analyses, and find that the models tend to utilize shallow lexical
hints \citep{shortcutlearning} in the target
language, rather than leveraging
abstract
crosslingual semantic features learned from the source language.

Our \textbf{contributions:}
\textbf{1)} We show that few-shot
crosslingual transfer is prone to large variations in task
performance; this property hinders unbiased assessments of the
effectiveness of different few-shot methods.
\textbf{2)} To remedy
this issue, we publish fixed and standardized few shots to support
fair comparisons and reproducibility.
\textbf{3)} We empirically
verify that few-shot crosslingual transfer has different performance
impact on structurally different tasks; we
provide in-depth
analyses concerning the source of performance gains.
\textbf{4)} We
analyze several SotA few-shot learning methods, and show that they
underperform simple full model finetuning. We hope that our work will
shed new light on the potential and current difficulties of few-shot
learning in crosslingual setups.

%% file: 03-related_work.tex
\section{Background and Related Work}

\textbf{Zero-/Few-Shot Crosslingual Transfer}. Multilingual pretrained 
encoders show strong zero-shot crosslingual transfer
(\textbf{\zsx}) ability in various NLP tasks
\citep{pires-etal-2019-multilingual, hsu-etal-2019-zero,
  artetxe-schwenk-2019-massively}.
In order to guide and measure the progress, standardized benchmarks like XTREME
\citep{hu2020xtreme} and XGLUE \citep{liang2020xglue} have been developed. 

Recently, \citet{lauscher2020zero} and
\citet{hedderich-etal-2020-transfer} extended the focus on
few-shot crosslingual transfer (\textbf{\fsx}): They assume the
availability of a handful of labeled examples in a target
language,\footnote{According to
  \newcite{garrette-baldridge-2013-learning}, it is possible to
  collect $\approx$100 POS-annotated sentences in two hours even for
  low-resource languages such as Malagasy.}
which are used to further finetune a source-trained model. The extra
few shots bring large performance gains at low annotation cost.
In this work, we systematically analyze this recent \fsx scenario.

\fsx resembles the intermediate-task transfer (STILT)
approach
\citep{phang2018sentence,pruksachatkun-etal-2020-intermediate}. In
STILT, a pretrained encoder is finetuned on a resource-rich intermediate
task, and then finetuned on a (resource-lean) target task.  Likewise,
\fsx focuses on transferring knowledge and general linguistic
intelligence \citep{yogatama2019learning}, although such transfer is
between \emph{languages} in the same task instead of between different
tasks.

\textbf{Few-shot learning} was first explored in computer vision 
\citep{miller2000learning, feifeilioneshot, Koch2015SiameseNN}; the
aim there is to learn new concepts with only few images.
Methods
like prototypical networks \citep{prototypicalnetsnell}
and model-agnostic meta-learning (MAML; \citet{pmlr-v70-finn17a})
have also been applied 
to many monolingual (typically English) NLP tasks
such as relation classification \citep{han-etal-2018-fewrel,gao-etal-2019-fewrel},
named-entity recognition \citep{DBLP:conf/acl/HouCLZLLL20},
word sense disambiguation \citep{holla-etal-2020-learning}, and
text classification \citep{yu-etal-2018-diverse,yin2020metalearning,
yin-etal-2020-universal,bansal-etal-2020-learning,gupta-etal-2020-effective}. 
However, recent few-shot learning methods in computer vision
consisting of two simple finetuning stages, first on base-class
images and then on new-class few shots, have been shown to outperform
MAML and achieve SotA scores
\citep{wang20j,chen2020new,eccvembeddingfewshot,Dhillon2020A}.
Inspired by this work, we compare various few-shot finetuning
methods from computer vision in the context of \fsx.

\textbf{Task Performance Variance}.
Deep neural networks' performance
on NLP tasks is bound to exhibit large variance.
\citet{reimers-gurevych-2017-reporting} and \citet{Dror:2019acl}
stress the importance of reporting \emph{score distributions}
instead of a single score for fair(er) comparisons.
\citet{dodge2020finetuning}, \citet{mosbach2020stability},
and \citet{zhang2020revisiting} show that finetuning pretrained
encoders with different random seeds yields performance with 
large variance.
In this work, we examine a specific source of variance: We show that
the choice of the few shots in crosslingual transfer learning also
introduces large variance in performance; consequently, we offer standardized few
shots for more controlled and fair comparisons.

%% file: 04-method.tex
\section{Method}

Following \citet{lauscher2020zero} and
\citet{hedderich-etal-2020-transfer}, our \fsx method comprises two
stages.
First, we conduct \textbf{source-training}: The pretrained
mBERT is finetuned with abundant annotated data in the source language.
Similar to \citet{hu2020xtreme}, \citet{liang2020xglue}
and due to the abundant labeled
data for many NLP tasks, we choose English as the source in our
experiments.
Directly evaluating the source-trained model after
this stage corresponds to the widely studied \zsx scenario.
The second
stage is \textbf{target-adapting}: The source-trained model from 
previous stage is adapted to a target language using few shots.
We discuss details of sampling the few shots in
\secref{datasetdiscussion}. The development set of the target language is used
for model selection in this stage.

%% file: 05-datasets_and_setup.tex
\section{Experimental Setup}
\seclabel{datasetdiscussion}
We consider three types of tasks requiring varying degrees of semantic
and syntactic knowledge transfer: Sequence classification
(\textbf{CLS}), named-entity recognition (\textbf{NER}), and
part-of-speech tagging (\textbf{POS}) in up to 40 typologically
diverse languages (cf., Appendix~\secref{appendixworkinglanguags}).

\subsection{Datasets and Selection of Few Shots}
For the CLS tasks, we sample few shots from four multilingual datasets:
News article classification (MLDoc; \citet{SCHWENK18.658});
Amazon review classification (MARC; \citet{keung-etal-2020-multilingual});
natural language inference (XNLI; \citet{conneau2018xnli,williams-etal-2018-broad}); and
crosslingual paraphrase adversaries
from word scrambling (PAWSX; \citet{zhang-etal-2019-paws,yang-etal-2019-paws}).
We use treebanks in Universal Dependencies
\citep{nivre-etal-2020-universal} for POS, and WikiANN dataset
\citep{pan-etal-2017-cross,rahimi-etal-2019-massively} for NER. 
\tabref{rawdatasetinfo} reports key information about the datasets.

\begin{table}[t]
\centering
\renewcommand{\arraystretch}{1.1}
\setlength{\tabcolsep}{2.5pt}
\resizebox{\linewidth}{!}{
{\scriptsize
\begin{tabularx}{\columnwidth}{cYcrYr}
\hline
Name  & Metric & Task                        & \multicolumn{1}{c}{$|\mathcal{T}|$} & TS & \multicolumn{1}{c}{\# of lang.} \\ \hline
XNLI  & Acc.   & Natural language inference  & 3            & No    & 15     \\
PAWSX & Acc.   & Paraphrase identification   & 2            & No    & 7        \\
MLDoc & Acc.   & News article classification & 4            & Yes   & 8        \\
MARC  & Acc.   & Amazon reviews              & 5            & Yes   & 6        \\ \hline
POS   & F1     & Part-of-speech tagging      & 17           & Yes   & 29  \\
NER   & F1     & Named-entity recognition    & 7            & Yes   & 40   \\ \hline
\end{tabularx}
}
}
\caption{Evaluation datasets. $|\mathcal{T}|$: Number of classes (classification tasks)
 and label set size (POS and NER). TS: availability of a training split in the target language.}
\tablabel{rawdatasetinfo}
\end{table}

We adopt the conventional few-shot sampling
strategy~\citep{feifeilioneshot, Koch2015SiameseNN,
  prototypicalnetsnell}, and conduct ``$N$-way $K$-shot'' sampling
from the datasets; $N$ is the number of classes and $K$ refers to the
number of shots per class.
A group of $N$-way $K$-shot data is referred to as a \textbf{bucket}. 
We set $N$ equal to the number of labels $|\mathcal{T}|$.
Following \citet{wang20j}, we sample 40 buckets for each target  (i.e., non-English)
language of a task to get a reliable estimation of model performance.

\textbf{CLS Tasks}.
For MLDoc and MARC, each language has a train/dev/test split.  We sample
the buckets without replacement from the training
set of each target language, so that buckets are disjoint from each other. 
Target languages in XNLI and PAWSX only have dev/test splits.  We
sample the buckets from the dev set; the remaining
data serves as a single new dev set for model selection during target-adapting.
For all tasks, we use $K \in \{1, 2, 4, 8\}$.

\textbf{POS and NER}.
For the two structured prediction tasks, 
``$N$-way $K$-shot'' is not well-defined because each
sentence contains one or more labeled tokens. 
We use a similar sampling principle as with CLS, 
where $N$ is the size of the label set for each language and task, 
but $K$ is set to the minimum number of occurrences for each label. 
In particular, we utilize the
\emph{Minimum-Including Algorithm} \citep{DBLP:journals/corr/abs-2009-08138,DBLP:conf/acl/HouCLZLLL20}
to satisfy the following criteria when 
sampling a bucket: 
1) each label appears at least $K$ times, and 
2) at least one label will appear less than $K$ times if any sentence is removed from the bucket.
Appendix~\secref{samplingalgorithm} gives sampling details.
In contrast to sampling for CLS, we do not enforce samples from
different buckets to be disjoint due to the  
small amount of data in some low-resource languages.
We only use $K \in \{1, 2, 4\}$ and exclude $K=8$, 
as 8-shot buckets already have lots of labeled tokens,
and thus (arguably) might not be considered few-shot.

\subsection{Training Setup} 
We use the pretrained cased mBERT model \citep{devlin-etal-2019-bert},
and rely on the PyTorch-based \cite{paszke2019pytorch} HuggingFace Transformers repository
\citep{Wolf2019HuggingFacesTS} in all experiments.

For \emph{source-training},
we finetune the pretrained encoder for 10 epochs with batch size 32. 
For \emph{target-adapting} to \emph{every} target language,
the few-shot data is a sampled bucket in this language, 
and we finetune on the bucket for 50 epochs with early-stopping of 10 epochs. 
The batch size is set to the number of shots in the bucket. 
Each target-adapting experiment is repeated 40 times using the 40 buckets.
We use the Adam  optimizer \citep{DBLP:journals/corr/KingmaB14}
with default
parameters in both stages with learning rates searched over $\{1e-5, 3e-5, 5e-5, 7e-5\}$. 
For CLS tasks, we use mBERT's \texttt{[CLS]} token as the final representation.
For NER and POS, following \citet{devlin-etal-2019-bert}, we use
a linear classifier layer on top of the representation of each tokenized
word, which is its last
wordpiece \citep{he2019establishing}.

We set the maximum sequence length to 128
after wordpiece tokenization \citep{wu2016googles}, in all
experiments.
Further implementation details
are shown in our Reproducibility Checklist in Appendix \secref{checklist}.

%% file: 06-discussion.tex
\section{Results and Discussion}
\subsection{Source-Training Results}
The \zsx performance from English (EN) to target languages of the four
CLS tasks are shown in the $K=0$ column in \tabref{bigclstable}.
For NER and POS, the results are shown in \figref{syntactictaskplots}.

For XTREME tasks (XNLI, PAWSX, NER, POS),
our implementation delivers results comparable to \citet{hu2020xtreme}.
For MLDoc, our results are comparable to
\citep{dong-de-melo-2019-robust,wu-dredze-2019-beto,eisenschlos-etal-2019-multifit}.
It is worth noting that reproducing the exact results is
challenging, as suggested by \citet{keung2020evaluation}.
For MARC, our zero-shot results are worse than
\citet{keung-etal-2020-multilingual}'s who use the dev set of
each target language for model selection while we use EN dev,
following the common true \zsx setup.

\begin{figure}[t]
  \centering
  \vspace{-1em}
  \subfloat{
    \includegraphics[width=.49\linewidth, trim={1.8cm 1.8cm 1.8cm 1.8cm}, clip]
    {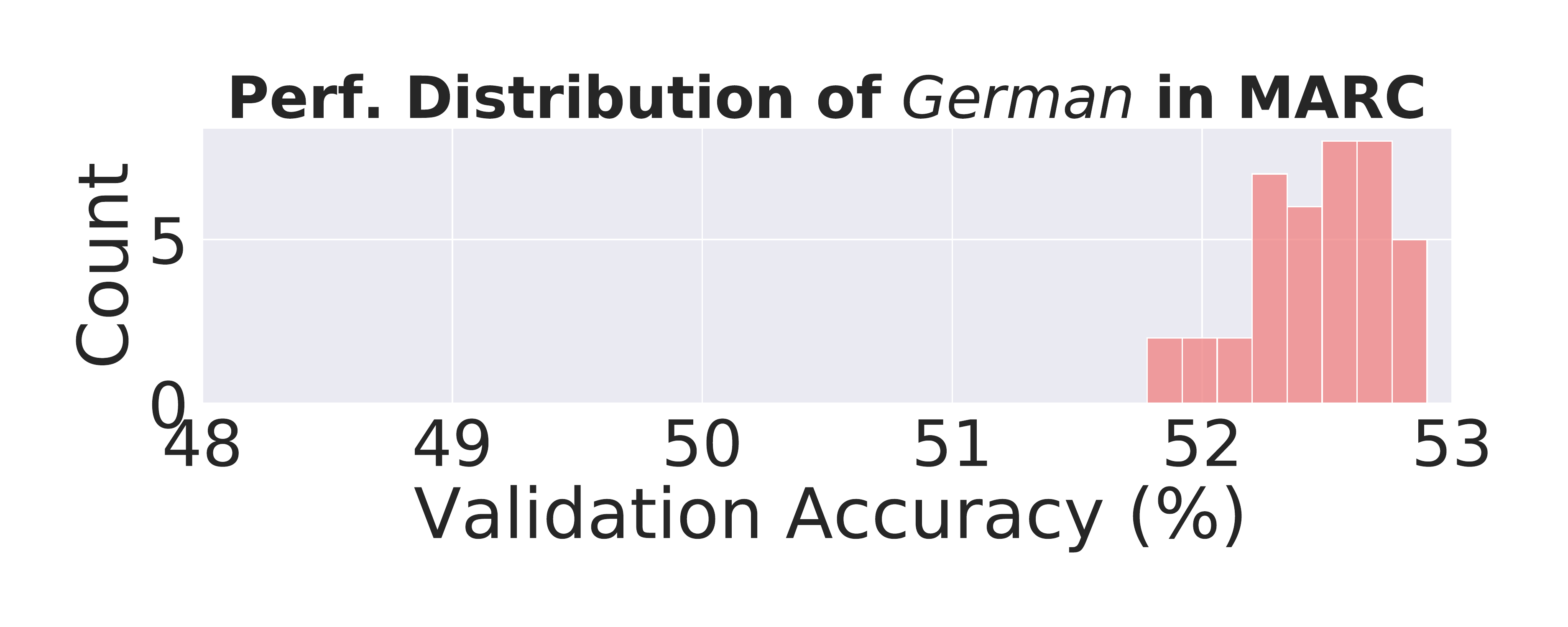}
  }
  \subfloat{
    \includegraphics[width=.49\linewidth, trim={1.8cm 1.8cm 1.8cm 1.8cm}, clip]
    {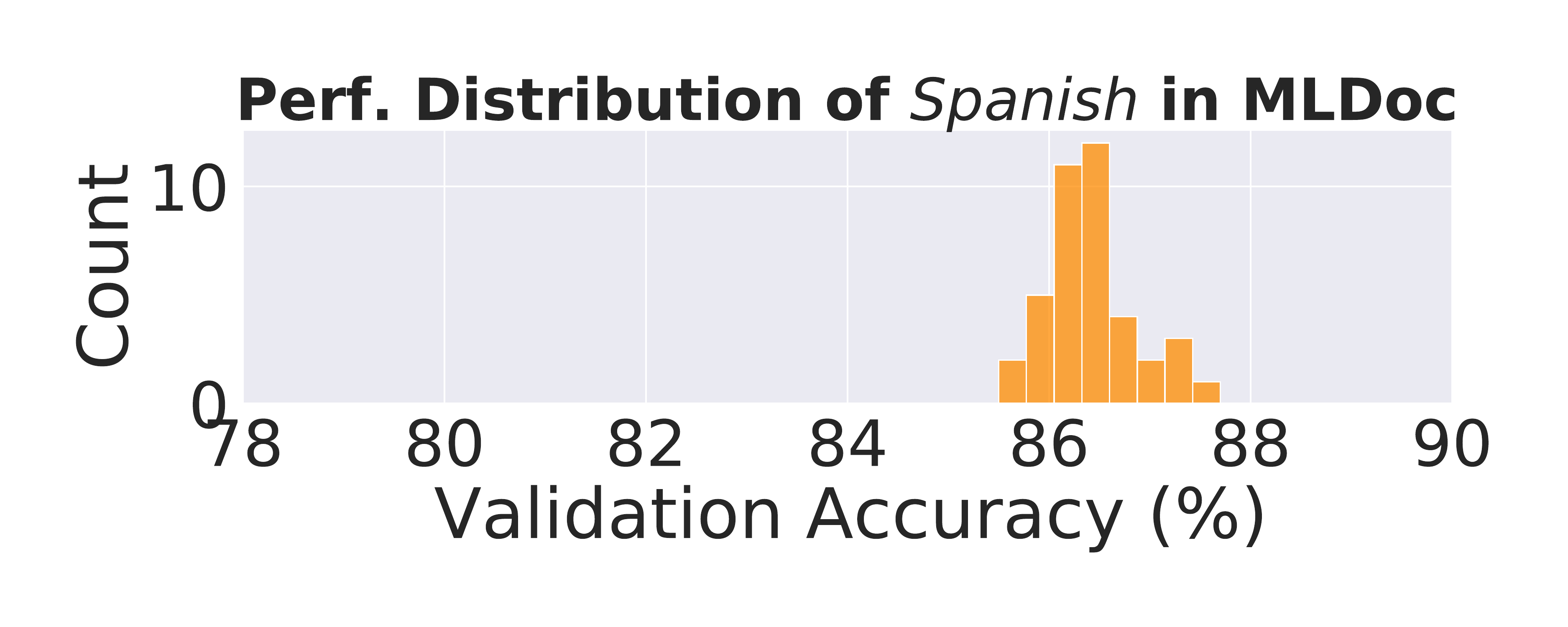}
  }
  \\
  \subfloat{
    \includegraphics[width=.49\linewidth, trim={1.8cm 1.8cm 1.8cm 1cm}, clip]
    {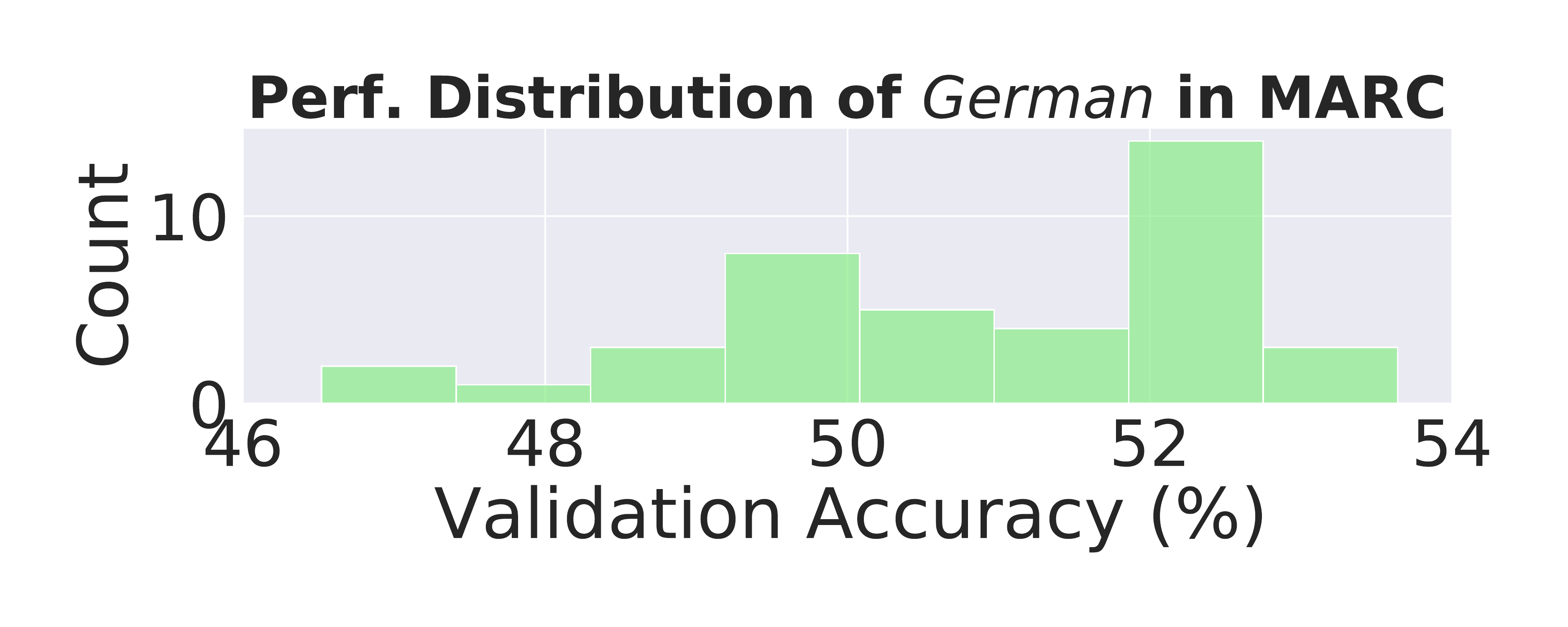}
  }
  \subfloat{
    \includegraphics[width=.49\linewidth, trim={1.8cm 1.8cm 1.8cm 1cm}, clip]
    {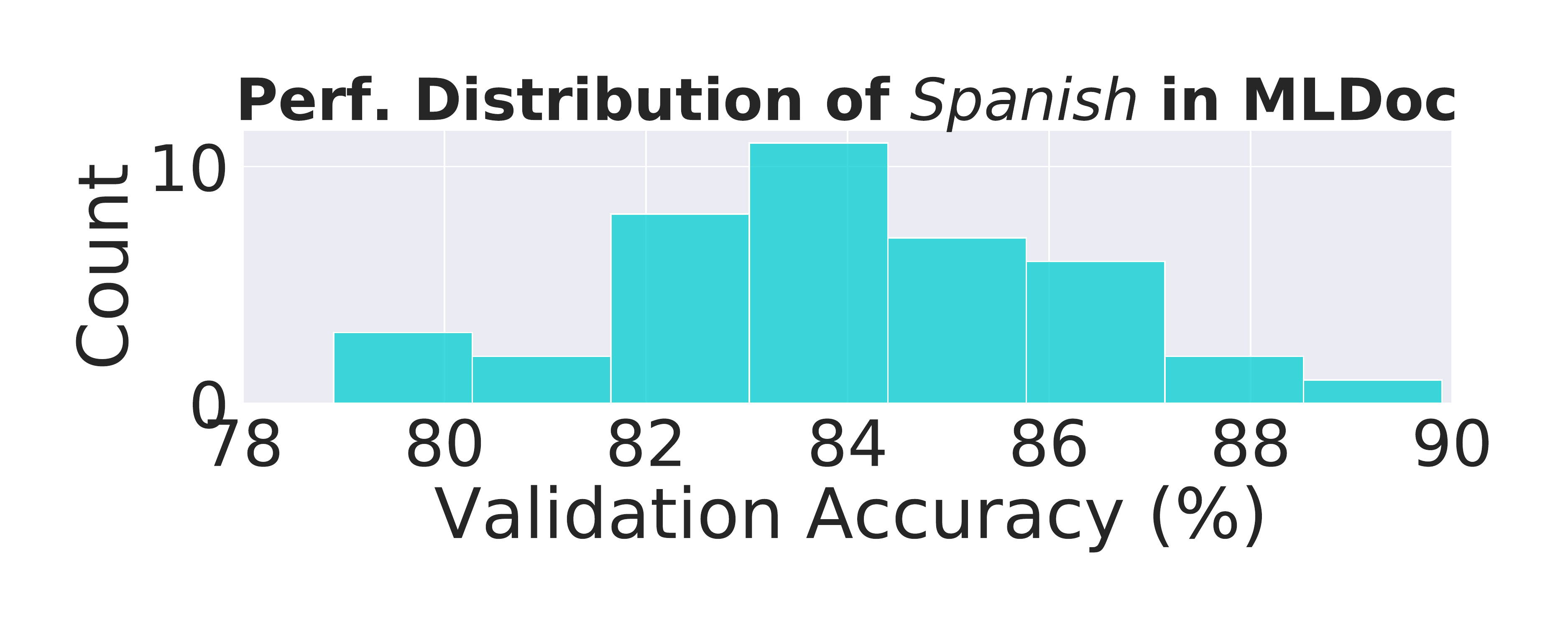}
  }
  \caption{Histograms of dev set accuracies.
Top: 40 runs with different random seeds. Bottom: 40 runs
with different 1-shot buckets. Left: DE MARC. Right:
ES MLDoc. The variance due to buckets is larger.}
  \figlabel{germanmarcdist}
  \vspace{-2mm}
\end{figure}

\subsection{Target-Adapting Results}
\textbf{Variance of Few-Shot Transfer}.
We hypothesize that \fsx suffers from large variance \citep{dodge2020finetuning}
due to the large model complexity and small amount of data in a  bucket.
To test this empirically, we first conduct two experiments on
MLDoc and MARC. First, for a \textit{fixed random seed}, we repeat 1-shot
target-adapting 40 times  using different 1-shot buckets in
German (DE) and Spanish (ES). Second, for a fixed 1-shot bucket, we repeat the same
experiment 40 times using random seeds in \{$0$ \ldots $39$\}.
\figref{germanmarcdist} presents the dev set performance distribution
of the 40 runs with 40 random seeds (top) and 40 1-shot buckets
(bottom).

With exactly the same training data, using different random seeds
yields a 1--2 accuracy difference of \fsx (\figref{germanmarcdist} top).
A similar phenomenon has been observed in finetuning monolingual
encoders \citep{dodge2020finetuning} and multilingual encoders with
\zsx \citep{keung2020evaluation,wu-dredze-2020-explicit,xia-etal-2020-bert};
we show this observation also holds for \fsx.
The key takeaway is that varying the buckets is a more severe
problem. It causes much larger variance (\figref{germanmarcdist}
bottom): The maximum accuracy difference is $\approx$6 for DE MARC and
$\approx$10 for ES MLDoc.
This can be due to the fact that difficulty of individual examples
varies in a dataset \citep{swayamdipta-etal-2020-dataset}, resulting
in different amounts of information encoded in buckets.

This large variance could be an issue when comparing different few-shot learning algorithms. 
The bucket choice is a strong confounding factor that may obscure the
strength of a promising few-shot technique.
Therefore, for fair comparison, \emph{it is necessary to work with a fixed set of few shots.}  
We propose to fix the sampled buckets for unbiased comparison of
different \fsx methods. We publish the sampled buckets from the six multilingual
datasets as a fixed and standardized few-shot evaluation benchmark.

In what follows, each \fsx
experiment is repeated 40 times using 40 different buckets with the same fixed random seed;
we report mean and standard deviation. 
As noted, the variance due to random seeds is 
smaller (cf., \figref{germanmarcdist}) and has been well
studied before \citep{reimers-gurevych-2017-reporting,dodge2020finetuning}. 
In this work, we thus focus our attention and limited
computing resources on understanding the impact of buckets, the \textit{newly detected}
source of variance.
However, we encourage practitioners to report results with both
factors considered in the future.

\input{bigclstable}

\textbf{Different Numbers of Shots}. A comparison concerning the number of shots ($K$), based 
on the few-shot results in \tabref{bigclstable} and 
\figref{syntactictaskplots}, reveals 
that the buckets largely improve model performance on a majority of
tasks (MLDoc, MARC, POS, NER) over 
zero-shot results. This is in line with
prior work \citep{lauscher2020zero,hedderich-etal-2020-transfer} and
follows the success of work on using bootstrapped 
data \citep{chaudhary-etal-2019-little,sherborne-etal-2020-bootstrapping}.

In general, we observe that: 
\textbf{1)} 1-shot buckets bring the largest relative performance
improvement over \zsx;
\textbf{2)} the gains follow the increase of $K$, but with diminishing returns; 
\textbf{3)} the performance variance across the 40 buckets
decreases as $K$ increases. 
These observations are more pronounced for POS and NER; e.g., 1-shot EN to Urdu (UR) POS transfer
shows gains of $\approx$22 $F_1$ points (52.40 with zero-shot, 74.95 with 1-shot).

For individual runs, we observe that models in \fsx tend to overfit the buckets quickly 
at small $K$ values.
For example, in around 32\% of NER 1-shot buckets, the model
achieves the best dev score right after the first
epoch; 
continuing the training only degrades performance.
Similar observations hold for semantic tasks like MARC, where in 10 out of 40
DE 1-shot buckets, the dev set performance peaks 
at epoch 1 (cf.\ learning curve in Appendix \secref{moreres} \figref{delearningcurve}).
This suggests the necessity of running the target-adapting 
experiments on multiple buckets if
reliable conclusions are to be drawn.

\def\permille{\ensuremath{{}^\text{o}\mkern-5mu/\mkern-3mu_\text{oo}}}
\begin{figure*}[t]
  \centering
  \subfloat{
    \includegraphics[width=\linewidth, trim={0.2cm 0.2cm 0cm 0.2cm}, clip]
    {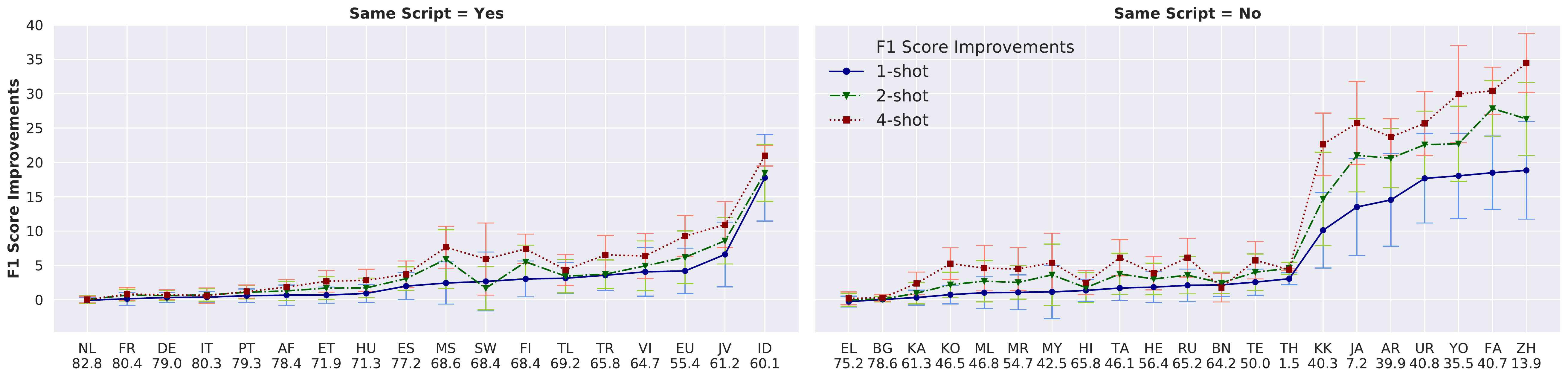}
  }
  \vspace{-1em}
  \subfloat{
    \includegraphics[width=\linewidth, trim={0.2cm 0.2cm 0cm 0.2cm}, clip]
    {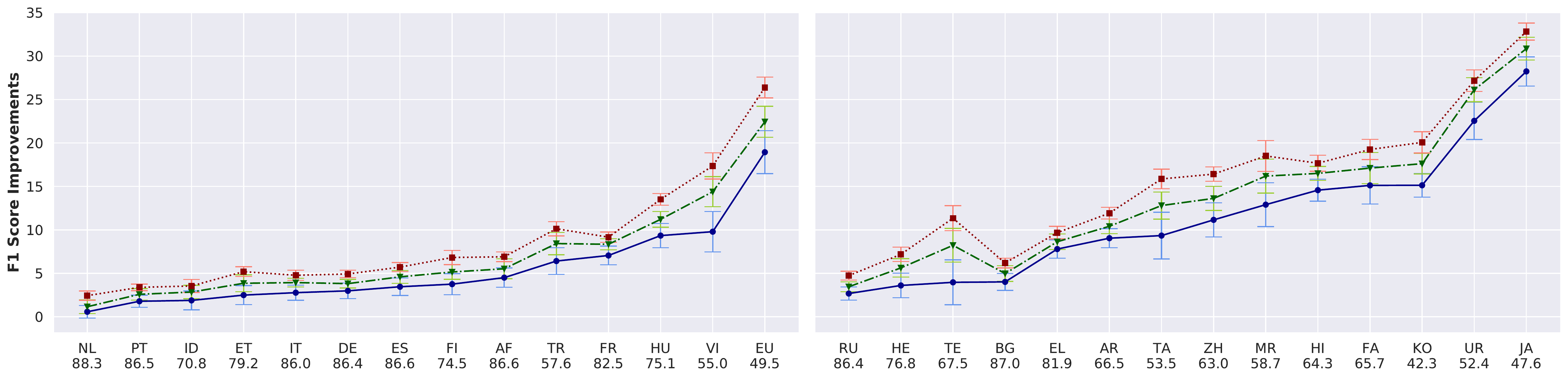}
  }
  \caption[Caption for LOF]{
  Improvement in $F_1$ (mean and standard deviation) of \fsx over \zsx 
    (numbers shown on x-axis beneath each language) for NER (top) and POS
    (bottom) for three different bucket sizes.
    See Appendix \secref{moreres} (\tabsref{posfewshot}
    and \tabrefbare{nerfewshot}) for absolute numerical values.  
}  
  \figlabel{syntactictaskplots}
\end{figure*}

\textbf{Different Downstream Tasks}.
The models for different tasks present various levels of sensitivity to \fsx. 
Among the CLS tasks that require semantic reasoning,
\fsx benefits
MLDoc the most.
This is not surprising given the fact that keyword matching can largely solve MLDoc
\citep{artetxe-etal-2020-cross,artetxe-etal-2020-call}: A few
examples related to target language keywords are expected to
significantly improve performance.
\fsx also yields prominent gains on the
Amazon review classification dataset MARC. 
Similar to MLDoc, we hypothesize that just matching a few important opinion and sentiment words
\citep{liu2012sentiment} in the target language brings large gains already. 
We provide further qualitative analyses in \secref{whyitworks}.

XNLI and PAWSX behave differently from
MLDoc and MARC. 
XNLI requires higher level semantic reasoning on pairs of sentences. 
\fsx performance improves modestly (XNLI) or even decreases (PAWSX-ES) compared 
to \zsx, even with large $K$. 
PAWSX requires a model to distinguish adversarially designed
non-paraphrase sentence pairs with large lexical overlap 
like ``Flights from New York to Florida''
and ``Flights from Florida to New York'' \citep{zhang-etal-2019-paws}. 
This poses a challenge for \fsx, given the small amount of target language information in the buckets. 
Therefore, \emph{when buckets are small (e.g., $K=1$) and
  for challenging semantic tasks like
  PAWSX, the buckets do not substantially help.}
Annotating more shots in the target language is an intuitive
solution.
Designing task-specific pretraining/finetuning objectives could also
be promising \citep{klein-nabi-2020-contrastive,ram2021few}.

Unlike CLS tasks,
POS and NER benefit from
\fsx substantially.
We speculate that there are
two reasons:
\textbf{1)} Both tasks often
require little to no high-level semantic understanding or reasoning;
\textbf{2)} due to i.i.d. sampling, train/dev/test splits are likely to have overlapping
vocabulary, and the labels in the buckets can easily propagate to 
dev and test. We delve deeper into these conjectures in \secref{whyitworks}.

\textbf{Different Languages}. 
For languages that are more distant from EN,
e.g., with different scripts, small lexical overlap, or fewer common typological features ~\citep{pires-etal-2019-multilingual,DBLP:conf/rep4nlp/WuD20},
\fsx introduces crucial lexical and structural information to 
guide the update of embedding and transformer layers in mBERT.

We present several findings based on the NER and POS results 
for a typologically diverse language sample. 
\figref{syntactictaskplots} shows that for languages with non-Latin scripts (different from EN), despite their
small to non-existent lexical overlap\footnote{We define lexical overlap as $\frac{|V|_{\textrm{L}} \cap |V|_{\textrm{\textsc{en}}}}{|V|_{\textrm{\textsc{en}}}}$ where $V$ denotes vocabulary. $|V|_{\textrm{L}}$ is computed with the 40  buckets of
  a target language L.\label{lo}}
and
diverging %
typological features
(see Appendix \secref{moreres} \tabsref{linguisticfeatures} and \tabrefbare{lexicaloverlap}),
the performance boosts are generally larger than those in the
same-script target languages: 6.2 vs.\ 3.0 average gain in NER and
11.4 vs.\ 5.4 in POS for $K=1$.
This clearly manifests the large
information discrepancy between target-language buckets and source-language data.
EN data is less relevant to these languages, so they obtain very
limited gain from source-training, reflected by their low \zsx scores.
With
a small amount of
target-language knowledge in the buckets, 
the performance is improved dramatically,
highlighting the effectiveness of \fsx.

\begin{table}[t]
\small\centering
\def\arraystretch{1.2}
\begin{tabular}{c|c|cc}
Task & Factor & S & P\\ \hline
\multirow{2}{*}{NER} & lexical overlap & -0.34 & -0.35\\
& \# of common linguistic features & -0.37 & -0.10\\ \hline
\multirow{2}{*}{POS} & lexical overlap   & -0.63 & -0.50\\
& \# of common linguistic features & -0.57 & -0.54\\ 
\end{tabular}
\caption{
  Correlations between \fsx $F_1$ score gains and
  the two
  factors (lexical overlap
  and the number of common linguistic features with EN)
  when considered
  independently for POS and NER: S/R denotes Spearman's/Pearson's $\rho$.
  See Footnotes \ref{lo}, \ref{lf} for information on the two factors.}
\tablabel{corrtable}
\end{table}

\tabref{corrtable} shows that, besides script form,
lexical overlap and the number of linguistic
features common with EN\footnote{Following
  \citet{pires-etal-2019-multilingual}, we use six
  WALS features: 81A (Order of Subject, Object and
  Verb), 85A (Order of Adposition and Noun), 86A (Order of
  Genitive and Noun), 87A (Order of Adjective and Noun), 88A
  (Order of Demonstrative and Noun), and 89A (Order of
  Numeral and Noun).\label{lf}}
also contribute directly to \fsx performance difference
among languages: There is
a moderate \textit{negative} correlation between $F_1$ score gains
vs.\ the two factors when considered independently
for
both syntactic tasks:
The fewer overlaps/features a target language
shares with EN, the larger the gain \fsx achieves.

This again stresses the importance of buckets %
-- they contain %
target-language-specific knowledge about a task that cannot be obtained by \zsx, which solely relies on language similarity.
Interestingly, Pearson's $\rho$ indicates that common linguistic features are much less linearly correlated with \fsx gains in NER than in POS.

\subsection{Importance of Source-Training}
\seclabel{scratchdiscussion}
\tabref{scratchtable} reports the performance \emph{drop} when directly carrying out
target-adapting, without any prior source-training of mBERT. 
We show the scores for
MLDoc and PAWSX
as a simple and a challenging
CLS task,
respectively.
For NER and POS, we select two high- (Russian (RU), ES), mid-
(Vietnamese (VI), Turkish (TR)), and low-resource languages (Tamil
(TA), Marathi (MR)) each.\footnote{The categorization based on resource
  availability is according to WikiSize
  \citep{DBLP:conf/rep4nlp/WuD20}.}

The results clearly indicate that
omitting the source-training stage yields large performance drops.
Even larger variance is also observed in this scenario (cf.\
Appendix~\secref{moreres} \tabref{absscratchtable}).
Therefore, the
model indeed learns, when trained on the source language, some
transferable crosslingual features that are beneficial to target
languages, both for semantic and syntactic tasks.

\begin{table}[t]
\scriptsize\centering
\renewcommand{\arraystretch}{1.2}
\setlength{\tabcolsep}{3pt}
\begin{tabular}{c|cc|cc|c|cc|cc}
 & \multicolumn{2}{c|}{MLDoc} & \multicolumn{2}{c|}{PAWSX} &  & \multicolumn{2}{c|}{POS} & \multicolumn{2}{c}{NER} \\
   & K=1     & K=8    & K=1    & K=8    &    & K=1    & K=4   & K=1     & K=4    \\ \hline
\textsc{de} & -37.73  & -7.67  & -31.11 & -30.82 & \textsc{ru} & -15.89 & -3.20 & -48.19  & -35.77 \\
\textsc{fr} & -38.14  & -13.21 & -33.02 & -32.34 & \textsc{es} & -9.51  & -0.93 & -63.98  & -41.53 \\
\textsc{es} & -33.69  & -14.38 & -33.76 & -33.97 & \textsc{vi} & -7.82  & -0.36 & -54.41  & -41.45 \\
\textsc{it} & -33.63  & -12.62 & -      & -      & \textsc{tr} & -15.05 & -8.08 & -54.35  & -34.52 \\
\textsc{ru} & -30.66  & -11.08 & -      & -      & \textsc{ta} & -13.72 & -4.40 & -34.70  & -24.81 \\
\textsc{zh} & -37.31  & -12.57 & -23.74 & -23.65 & \textsc{mr} & -11.34 & -3.63 & -40.10  & -25.68 \\
\textsc{ja} & -29.82  & -14.32 & -20.97 & -20.82 & -  & -      & -     & -       & -      \\
\textsc{ko} & -       & -      & -19.83 & -19.68 & -  & -      & -     & -       & -  
\end{tabular}
\caption{Performance drop when conducting target-adapting without
  source-training. 
 }
\tablabel{scratchtable}
\end{table}

\begin{figure}[t]
\centering
\includegraphics[width=.95\linewidth]{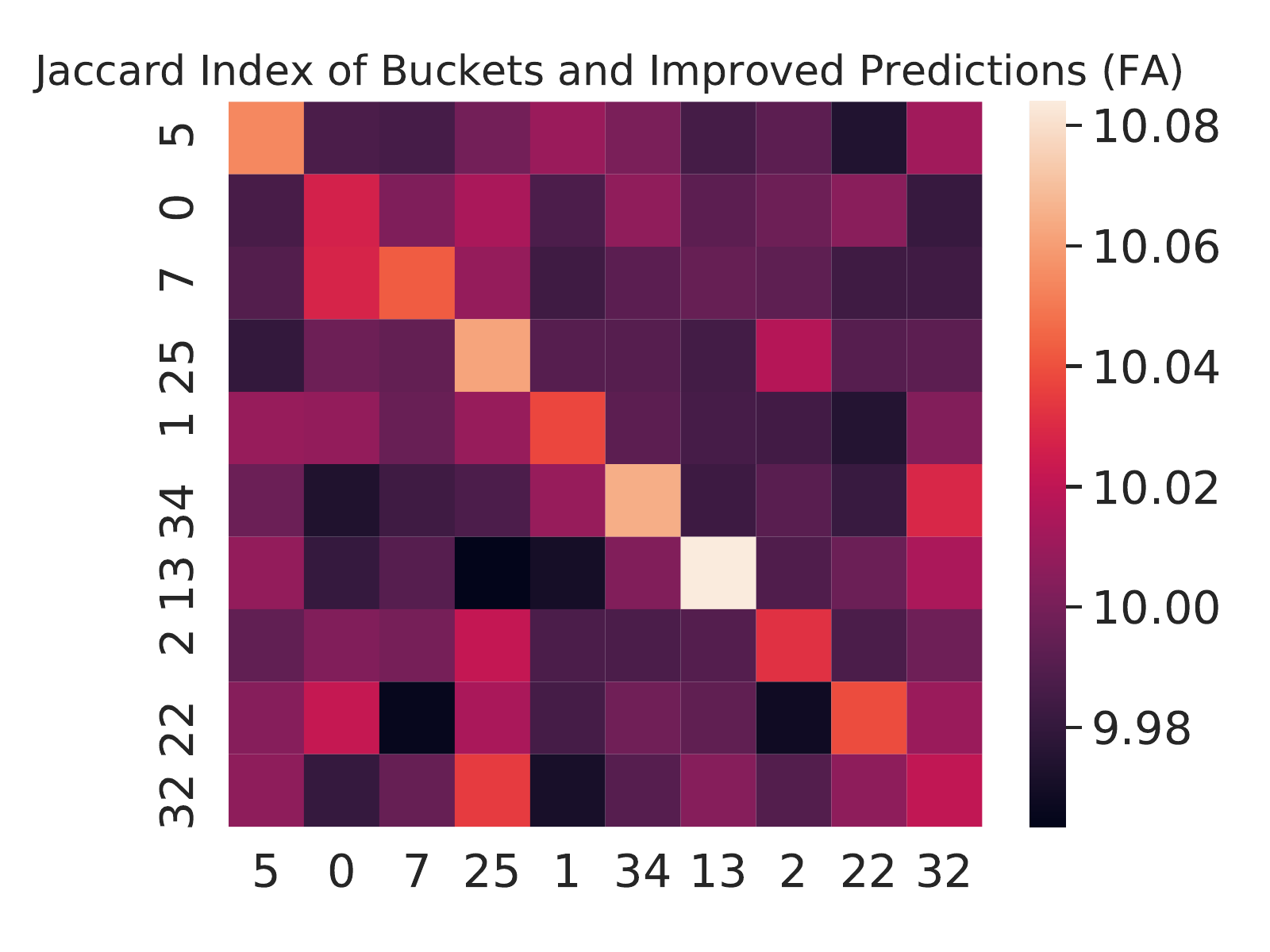}
\caption{Normalized (with softmax) Jaccard index (\%) of a bucket (row) and the
     improved predictions achieved with 10 buckets (column).
  }
\figlabel{jaccardpersian}
\end{figure}

\begin{figure}[t]
\centering
\includegraphics[width=.95\linewidth]{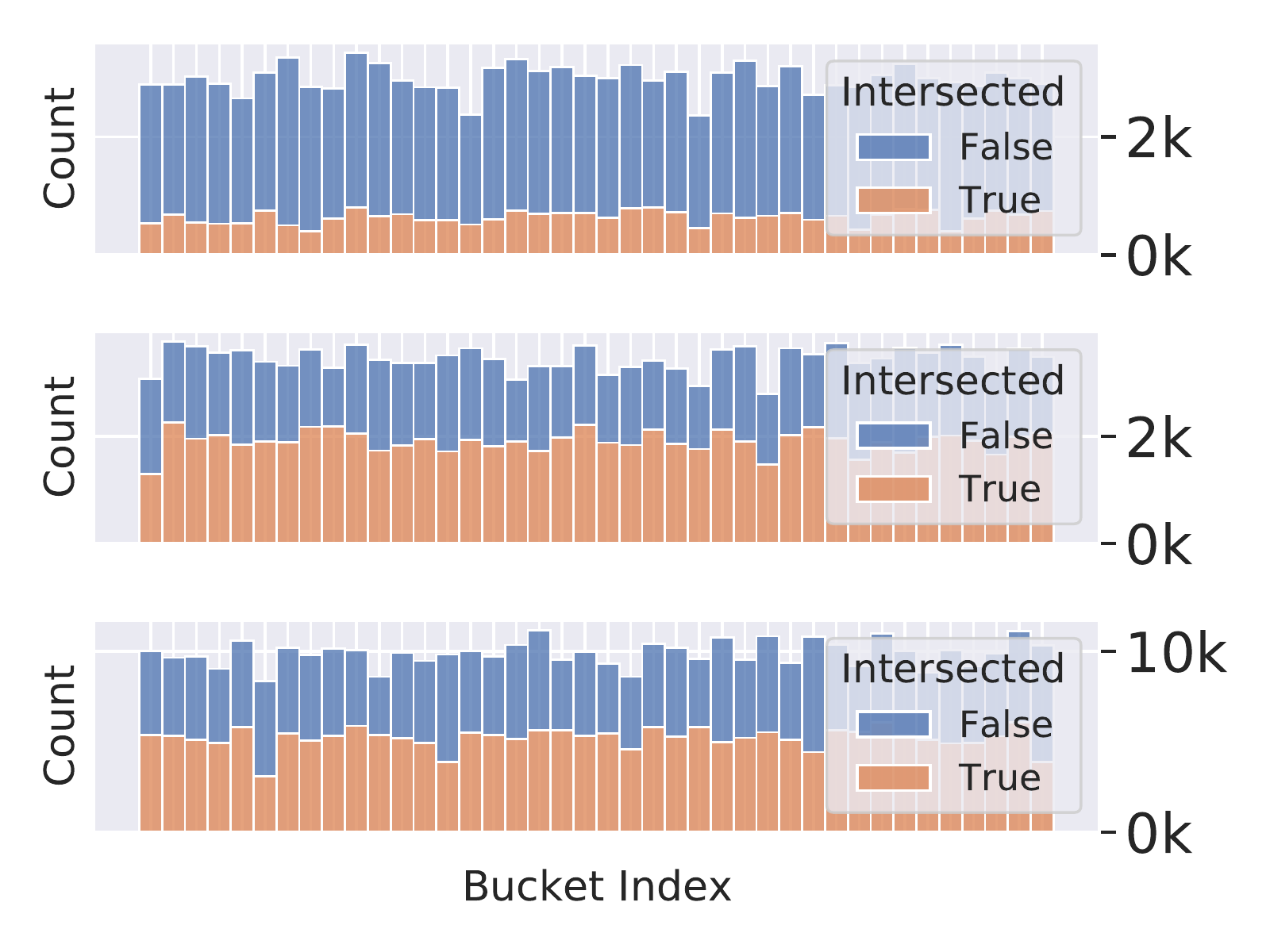}
\caption{
      Improvement of word-label predictions introduced by
      a bucket (x-axis) in FA (top), UR (mid),
      and HI (bottom), in relation to the words' presence in the bucket (True or False).
   }
\figlabel{traintestoverlap}
\end{figure}

\begin{figure}[t]
  \centering
  \subfloat{
    \includegraphics[width=.95\linewidth,height=.11875\textheight]
    {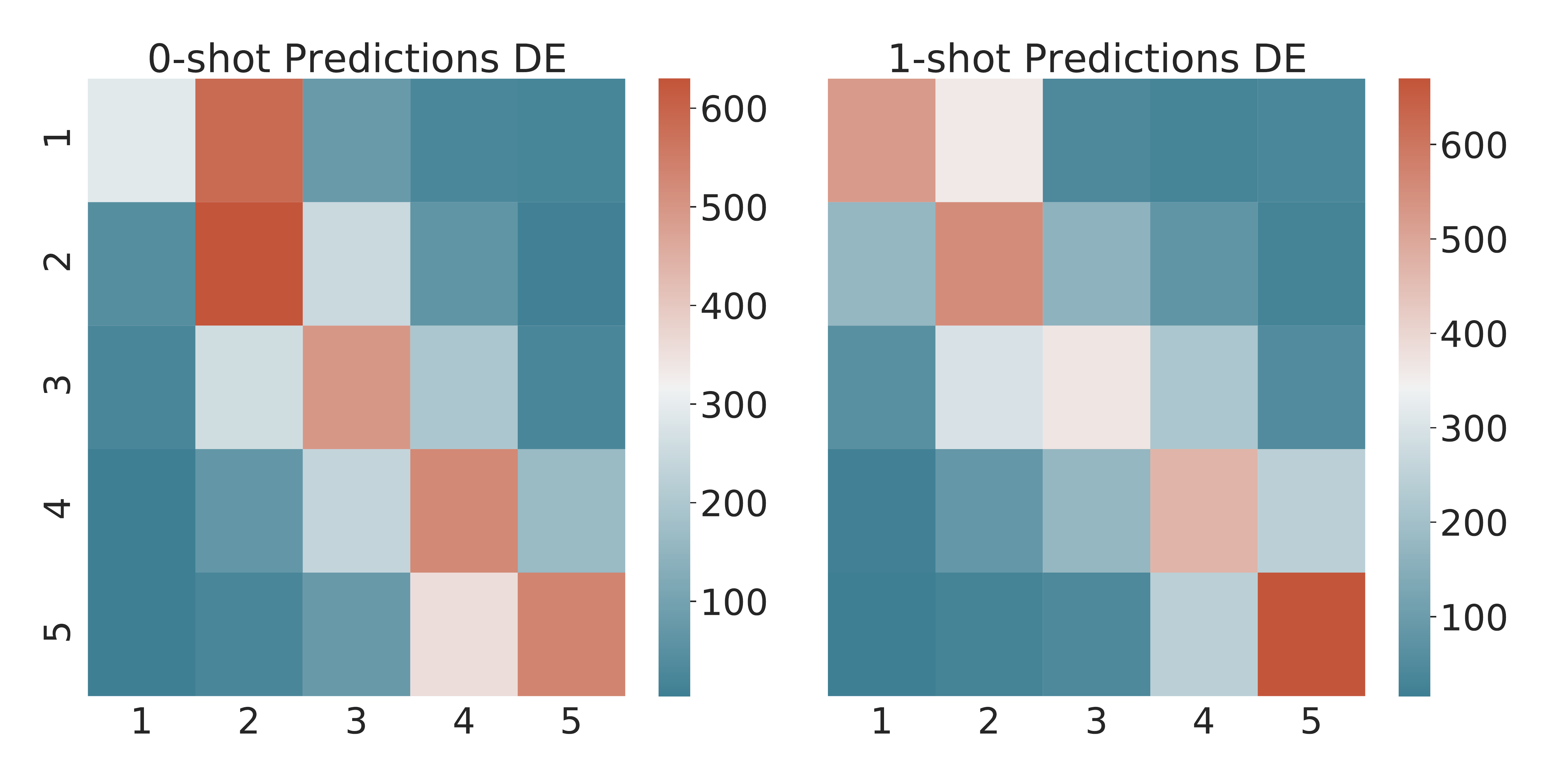}
  }
  \vspace{-1.7em}
  \subfloat{
    \includegraphics[width=.95\linewidth,height=.11875\textheight]
    {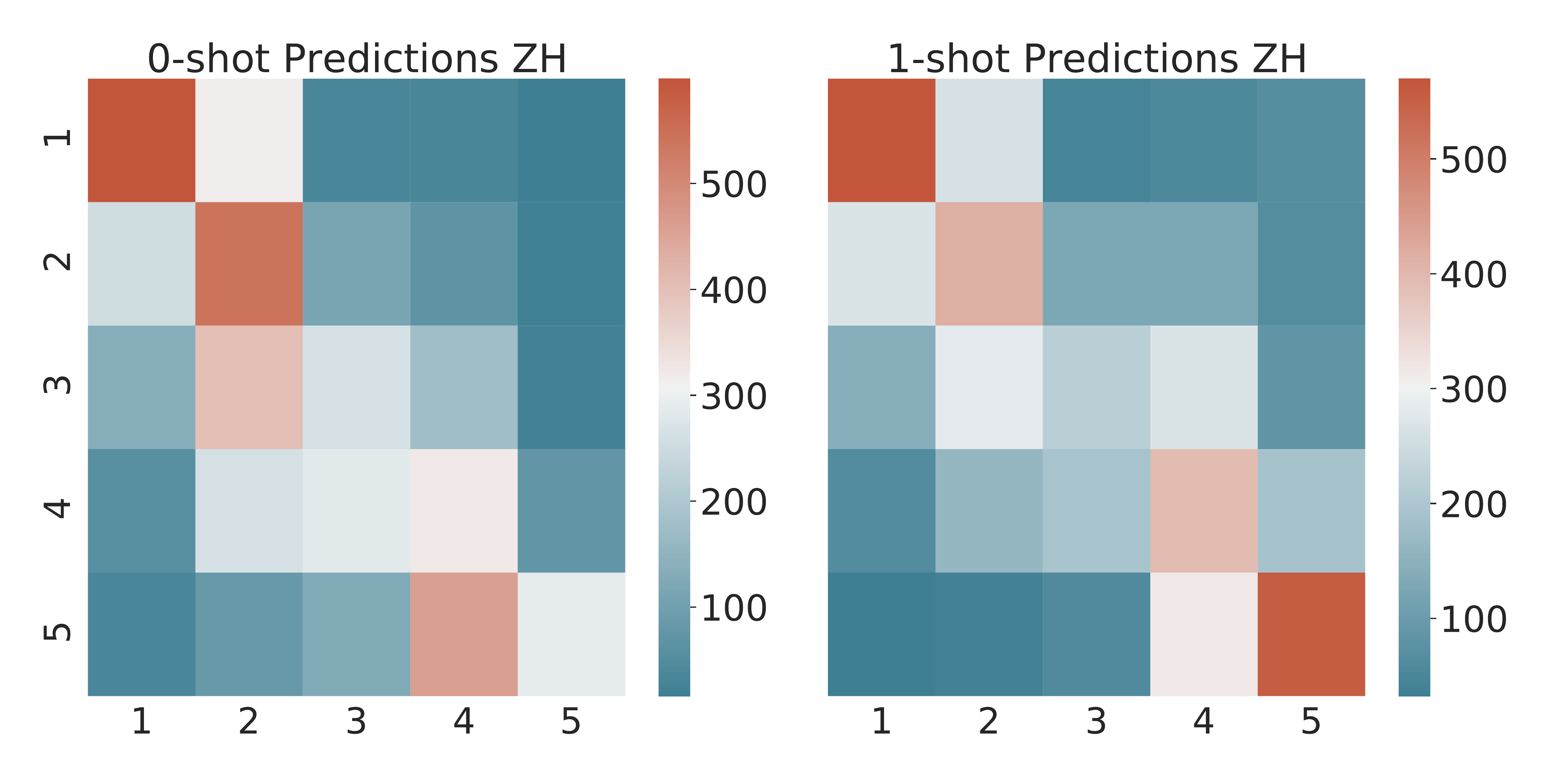}
  }
  \caption{
MARC (5 classes) test set prediction confusion matrices.
Top: DE. Bottom: ZH. Left: zero-shot models. Right:
1-shot models.  Colorbar numbers represent the
number of instances in that cell.
  }
  \figlabel{heatmap}
\end{figure}

\subsection{Importance of Lexical Features}
\seclabel{whyitworks}
We now investigate the sources of gains brought by
\fsx
over \zsx.

For syntactic tasks, we take Persian (FA) POS as an example.
\figref{jaccardpersian} visualizes the lexical overlap, measured by
the Jaccard index, of 10 1-shot buckets (rows) and the improved
word-label predictions introduced by target-adapting on each of the
buckets (columns).
In more detail,
for column $c$, we collect the set (denoted as $C_c$)
of all test set words whose
label is incorrectly predicted by the zero-shot model, but correctly
predicted by the model trained on the $c$-th bucket.
For row $i$, we denote with $B_i$  the set of words occurring in 
bucket $i$. 
The figure shows in cell ($i$, $k$) the Jaccard index of $B_i$ and $C_k$.
The bright color (i.e., higher lexical overlap) on the diagonal reflects that the improvements introduced
by a bucket are mainly\footnote{Note that the sampled
  buckets for POS are not completely disjoint (cf.\ sampling strategy in
  \secref{datasetdiscussion}).}  those
word-label predictions that are lexically more similar to the bucket
than to other buckets.

We also investigate the question: How many word-label predictions that are
improved after \fsx
occur in
the bucket, i.e., in the training data?
\figref{traintestoverlap} plots this for the 40 1-shot buckets in FA,
UR, and Hindi (HI).  We  see that many test words
do occur in the bucket~(shown in orange),
in line with recent findings
\citep{lewis-etal-2021-question,elangovan-etal-2021-memorization}.
These analyses shed light on why the buckets benefit NER/POS --
which heavily rely on lexical information -- more than higher level
semantic tasks.

For the CLS task MARC, which requires
understanding product reviews, \figref{heatmap}
visualizes the confusion matrices of  test set
predictions for DE and Chinese (ZH) zero-
and 1-shot models; axis ticks are
review scores in $\{1, 2, 3, 4, 5\}$. 
The squares on the diagonals in the two left heatmaps show that
parameter initialization on EN is a good basis for
well-performing \zsx:
This is
particularly true for DE, which is linguistically closer to
EN.  Two extreme review scores -- 1 (for DE) and 5 (for ZH)
-- have the largest confusions.
The two right heatmaps show that improvements
brought by the 1-shot buckets
are mainly achieved by correctly predicting more
cases of the two extreme review scores: 2 $\to$ 1 (DE) and 4 $\to$ 5 (ZH). 
But the more challenging cases
(reviews with scores 2, 3, 4), which require non-trivial reasoning, are
not significantly improved, or even become worse.

\begin{table}[t]
\scriptsize
\renewcommand{\arraystretch}{1.1}
\centering
\resizebox{1.0\columnwidth}{!}{
\begin{tabular}{c|cccccc}
token         & {[}SEP{]} & .     & nicht & !     & Die   & sehr    \\ \hline
$\Delta$Attn & +4.13     & +2.91 & +1.84 & -1.75 & -0.92 & -0.81
\end{tabular}
}%
\vspace{-1mm}
\caption{Tokens with the highest attention change from
  \texttt{[CLS]},
comparing zero-shot with
a  1-shot DE bucket.}
\tablabel{attnchange}
\end{table}

We inspect examples that are incorrectly predicted
by the few-shot model (predicting 1), 
but are correctly predicted by the zero-shot model (predicting 2).
Specifically, we compute the
difference of where \texttt{[CLS]} attends to, before and
after adapting the model on a 1-shot DE bucket.
We extract and average
attentions computed by the 12 heads from the topmost transformer
layer.  

\tabref{attnchange} shows that ``nicht'' (``not'') draws
high attention change from \texttt{[CLS]}.
``Nicht'' (i.e., negation) by itself is not a reliable
indicator of sentiment, so giving the lowest score to
reviews solely because they contain ``nicht'' is not a good strategy. 
The following review
is classified as 1 by the 1-shot model, but  2  is the
gold label (as the review is not entirely negative):

``Die Uhr ging nicht einmal eine Minute ...  \textbf{Optisch allerdings
  sehr schön}.'' \begin{small}(``The
    clock didn't even work one minute ...  \textbf{Visually, however, very
      nice}.'')\end{small}

Pretrained multilingual encoders are shown
to learn and store ``language-agnostic'' features \citep{pires-etal-2019-multilingual,zhao2020inducing};
\secref{scratchdiscussion}
shows that source-training mBERT on EN substantially benefits other
languages, even for difficult semantic tasks like
PAWSX.
Conditioning on
such language-agnostic features, we expect that
the buckets should lead
to good understanding and reasoning capabilities for a
target language.  However, plain few-shot
finetuning still relies heavily on
unintended shallow lexical
cues and shortcuts \citep{niven-kao-2019-probing,shortcutlearning}
that generalize poorly.
Other open
research questions for future work arise:
How do we overcome this excessive reliance on
lexical features? How can we leverage
language-agnostic features with \emph{few shots}?  
Our standardized buckets, baseline results,
and analyses are the initial step towards researching and answering these questions.

\input{clsconfigtable}

\subsection{Target-Adapting Methods}
\seclabel{variants}
SotA few-shot learning methods \citep{chen2018a,wang20j,eccvembeddingfewshot,Dhillon2020A}
from computer vision consist of 
two stages: 
1) training on base-class images, and
2) few-shot finetuning using new-class images.
Source-training and target-adapting stages of \fsx, albeit among languages, 
follow an approach
very similar to these methods.
Therefore, we test their effectiveness for crosslingual transfer.
These methods are built upon cosine similarity
that imparts inductive bias about distance and is 
more effective than
a fully-connected classifier 
layer (FC) with small $K$ \citep{wang20j}. %
Following
\citep{chen2018a,wang20j,eccvembeddingfewshot}, we freeze the
embedding and transformer layers of mBERT, and explore four
variants of the target-adapting stage using MARC.

\textbf{COS+Pooler}.
We randomly initialize a trainable weight matrix
$\mathbf{W} \in \mathbb{R} ^ {h \times c}$
where $h$ is the hidden
dimension size and  $c$ is the number of classes.
Rewriting $\mathbf{W}$ as
$[\mathbf{w}_{1},\ldots, \mathbf{w}_{i},\ldots,
\mathbf{w}_{c}]$,
we compute the logits of an input sentence representation
$\mathbf{x} \in \mathbb{R} ^ {h} $ (from mBERT) belonging to class $i$ as
$$\alpha \cdot \frac{\mathbf{x}^{\intercal}\mathbf{w}_i}{{\lVert \mathbf{x} \rVert_2 \cdot \lVert \mathbf{w}_i \rVert_2 }}\ ,$$
where $\alpha$ is a scaling hyperparameter, set to 10 in all experiments.
During 
training,
$\mathbf{W}$ and
mBERT's pooler layer  containing a linear layer and a \texttt{tanh} non-linearity
are updated.

\textbf{FC+Pooler}. During training, we update the linear classifier
layer and mBERT's pooler layer.

\textbf{FC only}. During training, we only update the linear
classifier layer.  This variant largely reduces model complexity and
exhibit lower variance when $K$ is small.

\textbf{FC(reset)+Pooler}. Similar to {FC+Pooler}, but the
source-trained linear classifier layer is randomly re-initialized before training.

\tabref{clsconfig} shows the performance of these methods along with
full model finetuning (without freezing).
\emph{FC+Pooler} performs the best among the four for both $K=1$ and
$K=8$ in all languages. However, it underperforms the full model
finetuning, especially when $K=8$.
\emph{FC only} is sub-optimal; yet
the decrease in comparison to \emph{FC+Pooler} is small, highlighting
that EN-trained mBERT is a strong feature extractor.
\emph{COS+Pooler} and \emph{FC(reset)+Pooler} perform considerably
worse than the other two methods and zero-shot transfer -- presumably
because their new parameters
need to be trained from scratch with few
shots.

We leave further exploration of other possibilities of exploiting
crosslingual features through collapse-preventing regularization
\citep{aghajanyan2021better} or contrastive learning
\citep{gunel2021supervised} to future work.
Integrating prompting \citep{gpt3,schick2020s,gao2020making,liu2021gpt} -- a
strong performing few-shot learning methodology for NLP -- into the
crosslingual transfer learning pipeline is also a promising direction.

%% file: bigclstable.tex
\newcommand{\WC}[1]{\cellcolor[HTML]{FFE2E0}{#1}}
\newcommand{\BC}[1]{\cellcolor[HTML]{D5FFD5}{#1}}

\definecolor{mygreen}{HTML}{D5FFD5}
\definecolor{myred}{HTML}{FFE2E0}

\begin{table}[t]
  \centering
  \tiny
  \renewcommand{\arraystretch}{1.1} 
  \setlength{\tabcolsep}{4pt}   
  \begin{tabular}{@{}cc|ccccc}
                   &    & K=0   & K=1              & K=2              & K=4              & K=8              \\ \hline
\multirow{8}{*}{\rotatebox[origin=c]{90}{\textbf{MLDoc}}} & \textsc{en} & 96.88 & -                & -                & -                & -                \\
                   & \textsc{de} & 88.30 & \BC{90.36 $\pm$ 1.48} & \BC{90.77 $\pm$ 0.87} & \BC{91.85 $\pm$ 0.83} & \BC{91.98 $\pm$ 0.82} \\
                   & \textsc{fr} & 83.05 & \BC{88.94 $\pm$ 2.46} & \BC{89.71 $\pm$ 1.68} & \BC{90.80 $\pm$ 0.88} & \BC{91.01 $\pm$ 0.94} \\
                   & \textsc{es} & 81.90 & \BC{83.99 $\pm$ 2.35} & \BC{85.65 $\pm$ 1.60} & \BC{86.30 $\pm$ 1.85} & \BC{88.46 $\pm$ 1.90} \\
                   & \textsc{it} & 74.13 & \BC{74.97 $\pm$ 2.04} & \BC{75.29 $\pm$ 1.57} & \BC{76.43 $\pm$ 1.41} & \BC{78.12 $\pm$ 1.25} \\
                   & \textsc{ru} & 72.33 & \BC{77.40 $\pm$ 4.27} & \BC{80.57 $\pm$ 1.37} & \BC{81.33 $\pm$ 1.33} & \BC{81.91 $\pm$ 1.21} \\
                   & \textsc{zh} & 84.38 & \BC{87.18 $\pm$ 1.45} & \BC{87.31 $\pm$ 1.53} & \BC{88.33 $\pm$ 1.11} & \BC{88.72 $\pm$ 1.05} \\
                   & \textsc{ja} & 74.58 & \BC{76.23 $\pm$ 1.59} & \BC{76.71 $\pm$ 2.12} & \BC{78.60 $\pm$ 2.43} & \BC{81.17 $\pm$ 1.72} \\ \hline\hline
\multirow{6}{*}{\rotatebox[origin=c]{90}{\textbf{MARC}}} & \textsc{en} & 64.52 & -                & -                & -                & -                \\
                   & \textsc{de} & 49.62 & \BC{51.50 $\pm$ 1.58} & \BC{52.76 $\pm$ 0.87} & \BC{52.78 $\pm$ 1.00} & \BC{53.32 $\pm$ 0.59} \\
                   & \textsc{fr} & 47.30 & \BC{49.32 $\pm$ 1.34} & \BC{49.70 $\pm$ 1.43} & \BC{50.64 $\pm$ 0.94} & \BC{51.23 $\pm$ 0.76} \\
                   & \textsc{es} & 48.44 & \BC{49.72 $\pm$ 1.24} & \BC{49.96 $\pm$ 1.12} & \BC{50.45 $\pm$ 1.22} & \BC{51.25 $\pm$ 0.93} \\
                   & \textsc{zh} & 40.40 & \BC{43.19 $\pm$ 1.76} & \BC{44.45 $\pm$ 1.36} & \BC{45.40 $\pm$ 1.26} & \BC{46.40 $\pm$ 0.93} \\
                   & \textsc{ja} & 38.84 & \BC{41.95 $\pm$ 2.09} & \BC{43.63 $\pm$ 1.30} & \BC{43.98 $\pm$ 0.89} & \BC{44.44 $\pm$ 0.69} \\ \hline\hline
\multirow{15}{*}{\rotatebox[origin=c]{90}{\textbf{XNLI}}} & \textsc{en} & 82.67 & -                &    -             &    -             &    -             \\
                   & \textsc{de} & 70.32 & \BC{70.58 $\pm$ 0.36} & \BC{70.60 $\pm$ 0.34} & \BC{70.61 $\pm$ 0.39} & \BC{70.70 $\pm$ 0.50} \\
                   & \textsc{fr} & 73.57 & \WC{73.41 $\pm$ 0.48} & \BC{73.74 $\pm$ 0.46} & \BC{73.57 $\pm$ 0.49} & \BC{73.77 $\pm$ 0.44} \\
                   & \textsc{es} & 73.71 & \BC{73.84 $\pm$ 0.40} & \BC{73.87 $\pm$ 0.44} & \BC{73.74 $\pm$ 0.48} & \BC{73.87 $\pm$ 0.46} \\
                   & \textsc{ru} & 68.70 & \BC{68.81 $\pm$ 0.52} & \BC{68.76 $\pm$ 0.54} & \BC{68.87 $\pm$ 0.55} & \BC{68.81 $\pm$ 0.77} \\
                   & \textsc{zh} & 69.32 & \BC{69.73 $\pm$ 0.94} & \BC{69.75 $\pm$ 0.94} & \BC{70.56 $\pm$ 0.76} & \BC{70.62 $\pm$ 0.86} \\
                   & \textsc{ar} & 64.97 & \WC{64.75 $\pm$ 0.36} & \WC{64.82 $\pm$ 0.23} & \WC{64.82 $\pm$ 0.23} & \WC{64.94 $\pm$ 0.37} \\
                   & \textsc{bg} & 67.58 & \BC{68.15 $\pm$ 0.69} & \BC{68.19 $\pm$ 0.75} & \BC{68.55 $\pm$ 0.67} & \BC{68.32 $\pm$ 0.70} \\
                   & \textsc{el} & 65.67 & \WC{65.64 $\pm$ 0.40} & \BC{65.73 $\pm$ 0.36} & \BC{65.80 $\pm$ 0.41} & \BC{66.00 $\pm$ 0.53} \\
                   & \textsc{hi} & 56.57 & \BC{56.94 $\pm$ 0.82} & \BC{57.07 $\pm$ 0.82} & \BC{57.21 $\pm$ 1.14} & \BC{57.82 $\pm$ 1.18} \\
                   & \textsc{sw} & 48.08 & \BC{50.33 $\pm$ 1.08} & \BC{50.28 $\pm$ 1.24} & \BC{51.08 $\pm$ 0.62} & \BC{51.01 $\pm$ 0.79} \\
                   & \textsc{th} & 46.17 & \BC{49.43 $\pm$ 2.60} & \BC{50.08 $\pm$ 2.42} & \BC{51.32 $\pm$ 2.07} & \BC{52.16 $\pm$ 2.43} \\
                   & \textsc{tr} & 60.40 & \BC{61.02 $\pm$ 0.68} & \BC{61.20 $\pm$ 0.61} & \BC{61.35 $\pm$ 0.49} & \BC{61.31 $\pm$ 0.56} \\
                   & \textsc{ur} & 57.05 & \BC{57.56 $\pm$ 0.85} & \BC{57.83 $\pm$ 0.91} & \BC{58.20 $\pm$ 0.93} & \BC{58.67 $\pm$ 1.03} \\
                   & \textsc{vi} & 69.82 & \BC{70.04 $\pm$ 0.59} & \BC{70.14 $\pm$ 0.75} & \BC{70.23 $\pm$ 0.63} & \BC{70.41 $\pm$ 0.70} \\ \hline\hline
\multirow{7}{*}{\rotatebox[origin=c]{90}{\textbf{PAWSX}}}   & \textsc{en} & 93.90 & -                & -                & -                & -                  \\
                  & \textsc{de} & 83.80 & \BC{84.14} $\pm$ 0.40 & \BC{84.08 $\pm$ 0.42} & \BC{84.04 $\pm$ 0.47} & \BC{84.23 $\pm$ 0.66}   \\
                  & \textsc{fr} & 86.90 & \BC{87.07} $\pm$ 0.27 & \BC{87.06 $\pm$ 0.37} & \BC{87.03 $\pm$ 0.31} & \BC{86.94 $\pm$ 0.41}   \\
                  & \textsc{es} & 88.25 & \WC{87.90} $\pm$ 0.54 & \WC{87.80 $\pm$ 0.56} & \WC{87.84 $\pm$ 0.53} & \WC{87.85 $\pm$ 0.75}   \\
                  & \textsc{zh} & 77.75 & \WC{77.71} $\pm$ 0.37 & \WC{77.63 $\pm$ 0.47} & \WC{77.68 $\pm$ 0.51} & \BC{77.82 $\pm$ 0.64}   \\
                  & \textsc{ja} & 73.30 & \BC{73.78} $\pm$ 0.75 & \BC{73.71 $\pm$ 1.04} & \BC{73.48 $\pm$ 0.69} & \BC{73.79 $\pm$ 1.28}   \\    
                  & \textsc{ko} & 72.05 & \BC{73.75} $\pm$ 1.30 & \BC{73.11 $\pm$ 1.05} & \BC{73.79 $\pm$ 0.92} & \BC{73.31 $\pm$ 0.61} \\

  \end{tabular}
  \caption{Zero-shot (column $K=0$) and few-shot (columns $K>0$) results (Acc. in \%)
    on the test set for CLS tasks. Green [red]: few-shot transfer outperforms [underperforms] zero-shot transfer.}
  \tablabel{bigclstable}
\end{table}

%% file: clsconfigtable.tex
\begin{table*}[t]
  \centering
  \tiny
  \setlength{\tabcolsep}{3.5pt}
  \renewcommand{\arraystretch}{1.2}
\begin{tabular}{@{}cc|cc|cc|cc|cc|cc|}
\cline{3-12}
 &
 &
  \multicolumn{2}{c|}{Full-Model Finetuning} &
  \multicolumn{2}{c|}{FC only} &
  \multicolumn{2}{c|}{FC + Pooler} &
  \multicolumn{2}{c|}{COS + Pooler} &
  \multicolumn{2}{c|}{FC (reset) + Pooler} \\ \cline{2-12} 
\multicolumn{1}{c|}{} &
  K=0 &
  K=1 &
  K=8 &
  K=1 &
  K=8 &
  K=1 &
  K=8 &
  K=1 &
  K=8 &
  K=1 &
  K=8 \\ \hline
\multicolumn{1}{|c|}{\textsc{de}} &
  49.62 &
  51.50 $\pm$ 1.58 &
  53.32 $\pm$ 0.59 &
  50.82 $\pm$ 1.17 &
  52.58 $\pm$ 0.63 &
  \textbf{51.18 $\pm$ 1.13} &
  \textbf{53.17 $\pm$ 0.58} &
  37.98 $\pm$ 5.53 &
  45.85 $\pm$ 2.14 &
  38.52 $\pm$ 6.64 &
  49.46 $\pm$ 2.21 \\
\multicolumn{1}{|c|}{\textsc{fr}} &
  47.30 &
  49.32 $\pm$ 1.34 &
  51.23 $\pm$ 0.76 &
  48.19 $\pm$ 0.78 &
  49.05 $\pm$ 0.93 &
  \textbf{48.60 $\pm$ 1.02} &
  \textbf{49.97 $\pm$ 0.77} &
  39.93 $\pm$ 3.50 &
  44.41 $\pm$ 1.95 &
  40.12 $\pm$ 5.04 &
  47.77 $\pm$ 2.00 \\
\multicolumn{1}{|c|}{\textsc{es}} &
  48.44 &
  49.72 $\pm$ 1.24 &
  51.25 $\pm$ 0.93 &
  49.03 $\pm$ 0.73 &
  49.69 $\pm$ 0.57 &
  \textbf{49.28 $\pm$ 0.85} &
  \textbf{50.21 $\pm$ 0.63} &
  40.01 $\pm$ 4.33 &
  45.35 $\pm$ 2.37 &
  40.89 $\pm$ 4.96 &
  47.73 $\pm$ 2.33 \\
\multicolumn{1}{|c|}{\textsc{zh}} &
  40.40 &
  43.19 $\pm$ 1.76 &
  46.40 $\pm$ 0.93 &
  41.90 $\pm$ 1.15 &
  43.34 $\pm$ 0.88 &
  \textbf{42.30 $\pm$ 1.37} &
  \textbf{44.42 $\pm$ 0.65} &
  33.10 $\pm$ 5.48 &
  38.31 $\pm$ 1.87 &
  31.83 $\pm$ 7.00 &
  42.07 $\pm$ 2.19 \\
\multicolumn{1}{|c|}{\textsc{ja}} &
  38.84 &
  41.95 $\pm$ 2.09 &
  44.44 $\pm$ 0.69 &
  40.76 $\pm$ 1.76 &
  43.14 $\pm$ 0.76 &
  \textbf{41.40 $\pm$ 1.74} &
  \textbf{43.81 $\pm$ 0.56} &
  34.36 $\pm$ 4.19 &
  38.95 $\pm$ 1.80 &
  32.80 $\pm$ 5.17 &
  41.18 $\pm$ 1.68 \\ \hline
\end{tabular}
\caption{
  Accuracy (\%) on MARC when varying classifier head configurations.
  Full-Model Finetuning updates all parameters during
  training; the other four methods only update a
  subset as described in \secref{variants}.
  The best results (excluding Full-Model Finetuning) are in bold.
}
\tablabel{clsconfig}
\end{table*}

%% file: 07-conclusion.tex
\section{Conclusion and Future Work}
We have presented an extensive study 
of \emph{few-shot crosslingual transfer}. 
The focus of the study has been on an empirically detected performance
variance in few-shot scenarios: The models exhibit a high level of
sensitivity to the choice of few shots. We analyzed and discussed the
major causes of this variance across six diverse tasks
for up to 40 languages.
Our results show that large language models
tend to overfit to few shots quickly and mostly rely on shallow
lexical features present in the few shots, though they have been
trained with abundant data in English.
Moreover, we have
empirically validated that
state-of-the-art
few-shot learning methods in computer vision do not outperform a
conceptually simple alternative: Full model finetuning.

Our study calls for more rigor and accurate reporting of the results
of few-shot crosslingual transfer experiments.  They should include
score distributions over standardized and fixed few shots.  To aid
this goal, we have created and provided such fixed few shots as a
standardized benchmark for six multilingual datasets.

Few-shot learning is promising for crosslingual transfer, because it
mirrors how people acquire new languages, and that the few-shot
data annotation is feasible.  In future work, we will investigate more
sophisticated techniques and extend the work to more NLP tasks.

%% file: 99-appendix.tex
\section{Reproducibility Checklist}
\seclabel{checklist}

\subsection{mBERT Architecture and Number of Parameters}
\seclabel{bertdetails}

We use the ``bert-base-multilingual-cased''
model\footnote{\url{https://github.com/google-research/bert/blob/master/multilingual.md}}.
It contains 12 Transformer blocks with 768 hidden dimensions. Each
block has 12 self attention heads. The model is pretrained on the
concatenation of the Wikipedia dump of 104 languages.

There are about 179 million parameters in mBERT.
For all the tasks, we use a linear output layer.
Denoting the output dimension of a task as $m$, e.g., $m=2$ for PAWSX.
Then we have in total 179 million + 768$\times m$ + $m$ parameters for the task.

\subsection{Computing Infrastructure}
All experiments are conducted on GeForce GTX 1080Ti.  In the
source-training stage, we use 4 GPUs with per-GPU batch size 32. In
the target-adapting stage, we use a single GPU and the batch size is
equal to the number of examples in a bucket.

\subsection{Evaluation Metrics and Validation Performance}
We follow the standard evaluation metrics used in XTREME
\citep{hu2020xtreme} and they are shown in \tabref{rawdatasetinfo};
evaluation
functions in \texttt{scikit-learn} \citep{scikit-learn} and \texttt{seqeval}
(\url{https://github.com/chakki-works/seqeval}) are used.
Link to code: \texttt{code/utils/eval\_meters.py}.

The validation performance of the English-trained models are shown in
the first row of \tabref{envalperf}; the optimal learning rate for
each task is shown in the second row.

\begin{table}[h!]
\centering
\scriptsize
\renewcommand{\arraystretch}{1.2}
\begin{tabular}{cccccc}
MLDoc  & MARC  & XNLI  & PAWSX & POS  & NER  \\ \hline
  98.1 & 65.1  & 83.5  & 94.5  & 95.6 & 84.3 \\
 1e-5  & 1e-5  & 3e-5 & 1e-5  & 1e-5 & 1e-5                   
\end{tabular}
\caption{Source-training validation performance (\%) and the optimal learning rate.}
\tablabel{envalperf}
\end{table}

For all the \fsx experiments, we enclosed the
validation scores
in
\url{https://github.com/fsxlt/running-logs}.

\subsection{Hyperparameter Search}
For both source-training and target-adapting, the only hyperparameter 
we search is
learning rate (from $\{1e-5, 3e-5, 5e-5, 7e-5\}$)
to reduce the
sensitivity of our results to hyperparameter selection.

\subsection{Datasets and Preprocessing}
For tasks (XNLI, PAWSX, POS, NER) covered in XTREME \citep{hu2020xtreme},
we utilize the provided preprocessed datasets.
Our MLDoc dataset is obtained from
\url{https://github.com/facebookresearch/MLDoc}. 
We retrieve MARC  from
\url{docs.opendata.aws/amazon-reviews-ml/readme.html}.
\tabref{datasetexamples} shows example entries of the datasets.
It is worth noting that MARC is a single sentence review
classification task, however, we put the ``review title'' and  ``product category''
in the ``Text B'' field, following \citet{keung-etal-2020-multilingual}.

We utilize the tokenizer in the HuggingFace \texttt{Transformers}
package \citep{Wolf2019HuggingFacesTS} to preprocess all the texts.
In all experiments, we use 128 maximum sequence length and truncate from
the end of a sentence if its length exceeds the limit.

\input{dataexampletableappendix}

\section{Languages}
\seclabel{appendixworkinglanguags}
We work on 40 languages in total.
They are shown in \tabref{linguisticfeatures}, together with their ISO 639-1 codes, writing script, and language features from WALs (\url{https://wals.info/}) used in our experiments.

\input{linguistictable}

\section{Minimum-Including Algorithm}
\seclabel{samplingalgorithm}
\input{minincl}

We utilize the \emph{Minimum-including Algorithm} from
\citet{DBLP:conf/acl/HouCLZLLL20,DBLP:journals/corr/abs-2009-08138}
for sampling the buckets of
POS and NER which have several labels in a sentence.
Denoting as $\bm{x}$ a sentence that consists of an array of words
($x_1,\ldots,x_n$), and the  array $\bm{y}$ that consists of a
series of labels ($y_1,\ldots,y_n$). We sample the buckets by using
Algorithm \ref{miniincludealgorithm}. Note that we sample with
replacement for POS and NER.

\section{Additional Results}
\seclabel{moreres}

\subsection{Learning Curve}
\figref{delearningcurve} visualizes the averaged learning curve of 10 out of 40 German 1-shot MARC
buckets for which the best dev performance is obtained at epoch 1.

\subsection{Numerical Values}
The numerical values of the POS and NER \fsx results
are shown in \tabref{nerfewshot} and \tabref{posfewshot}.
The absolute performances of few-shot transfer without English source-training
are shown in \tabref{absscratchtable}.
The lexical overlap of target languages with EN for NER and POS is shown in \tabref{lexicaloverlap}.

\begin{table}[t]
\centering\tiny\renewcommand{\arraystretch}{1.2}
\begin{tabular}{|ccccc|ccccc|}
\hline
292 & 584 & 78  & 27  & 19  & 526 & 361 & 43  & 31  & 40  \\
45  & 630 & 250 & 64  & 11  & 176 & 554 & 162 & 80  & 28  \\
24  & 259 & 497 & 196 & 24  & 65  & 298 & 369 & 218 & 50  \\
4   & 69  & 237 & 525 & 165 & 22  & 87  & 176 & 471 & 244 \\
6   & 25  & 75  & 357 & 537 & 16  & 27  & 42  & 245 & 670 \\ \hline
599 & 316 & 36  & 33  & 16  & 570 & 262 & 45  & 56  & 67  \\
255 & 543 & 112 & 70  & 20  & 269 & 416 & 126 & 125 & 65  \\
136 & 401 & 266 & 174 & 23  & 143 & 284 & 219 & 270 & 84  \\
60  & 262 & 283 & 322 & 73  & 63  & 163 & 190 & 395 & 189 \\
38  & 83  & 127 & 462 & 290 & 32  & 39  & 59  & 314 & 555 \\ \hline
\end{tabular}
\caption{Numerical value of the confusion matrices in
  \figref{heatmap}. For 1-shot confusion matrices (right), we average
results of 5 buckets and then round to integers.}
\end{table}

\begin{table*}[!]
\centering
\tiny
\def\arraystretch{0.93} 
\begin{tabular}{c|cc|cc|c|cc|cc}
 & \multicolumn{2}{c|}{MLDoc} & \multicolumn{2}{c|}{PAWSX} &  & \multicolumn{2}{c|}{POS} & \multicolumn{2}{c}{NER} \\
   & K=1              & K=8              & K=1              & K=8               &    & K=1              & K=4              & K=1              & K=4             \\ \hline
DE & 52.63 $\pm$ 8.98 & 84.31 $\pm$ 3.60 & 53.03 $\pm$ 1.67 & 53.41 $\pm$ 1.47  & RU & 73.18 $\pm$ 4.42 & 86.65 $\pm$ 1.32 & 19.11 $\pm$ 6.94 & 35.57 $\pm$ 6.23 \\
FR & 50.80 $\pm$ 8.50 & 77.80 $\pm$ 4.44 & 54.05 $\pm$ 1.33 & 54.60 $\pm$ 0.97  & ES & 80.54 $\pm$ 4.17 & 90.26 $\pm$ 0.99 & 15.21 $\pm$ 5.98 & 39.37 $\pm$ 5.33 \\
ES & 50.30 $\pm$ 8.30 & 74.08 $\pm$ 6.48 & 54.14 $\pm$ 1.53 & 53.88 $\pm$ 1.72  & VI & 56.97 $\pm$ 5.16 & 72.00 $\pm$ 1.99 & 14.36 $\pm$ 4.28 & 29.63 $\pm$ 5.55\\
IT & 41.34 $\pm$ 6.82 & 65.50 $\pm$ 4.21 & -                & -                 & TR & 48.96 $\pm$ 3.15 & 59.65 $\pm$ 1.83 & 15.02 $\pm$ 5.58 & 37.81 $\pm$ 5.63 \\
RU & 46.74 $\pm$ 9.48 & 70.83 $\pm$ 5.63 & -                & -                 & TA & 49.12 $\pm$ 4.67 & 64.96 $\pm$ 2.16 & 13.11 $\pm$ 4.55 & 27.42 $\pm$ 4.82   \\
ZH & 49.87 $\pm$ 10.44& 76.15 $\pm$ 5.10 & 53.97 $\pm$ 1.79 & 54.17 $\pm$ 1.38  & MR & 60.26 $\pm$ 5.72 & 73.58 $\pm$ 2.39 & 15.68 $\pm$ 7.09 & 33.50 $\pm$ 6.02   \\
JA & 46.41 $\pm$ 6.59 & 66.85 $\pm$ 6.54 & 52.81 $\pm$ 0.96 & 52.97 $\pm$ 1.15  & -  & -                & -                & -         &     -         \\    
KO & -                & -                & 53.92 $\pm$ 0.78 & 53.63 $\pm$ 0.99  & -  & -                & -                & -         &    -

\end{tabular}
\caption{Target-adapting results without
  source-training. Numbers are mean and standard deviation of 40
  runs.}
\tablabel{absscratchtable}
\end{table*}

\input{posfewshottable}

\input{nerfewshottable}

\begin{figure}[h!]
  \centering
  \includegraphics[width=.8\linewidth]{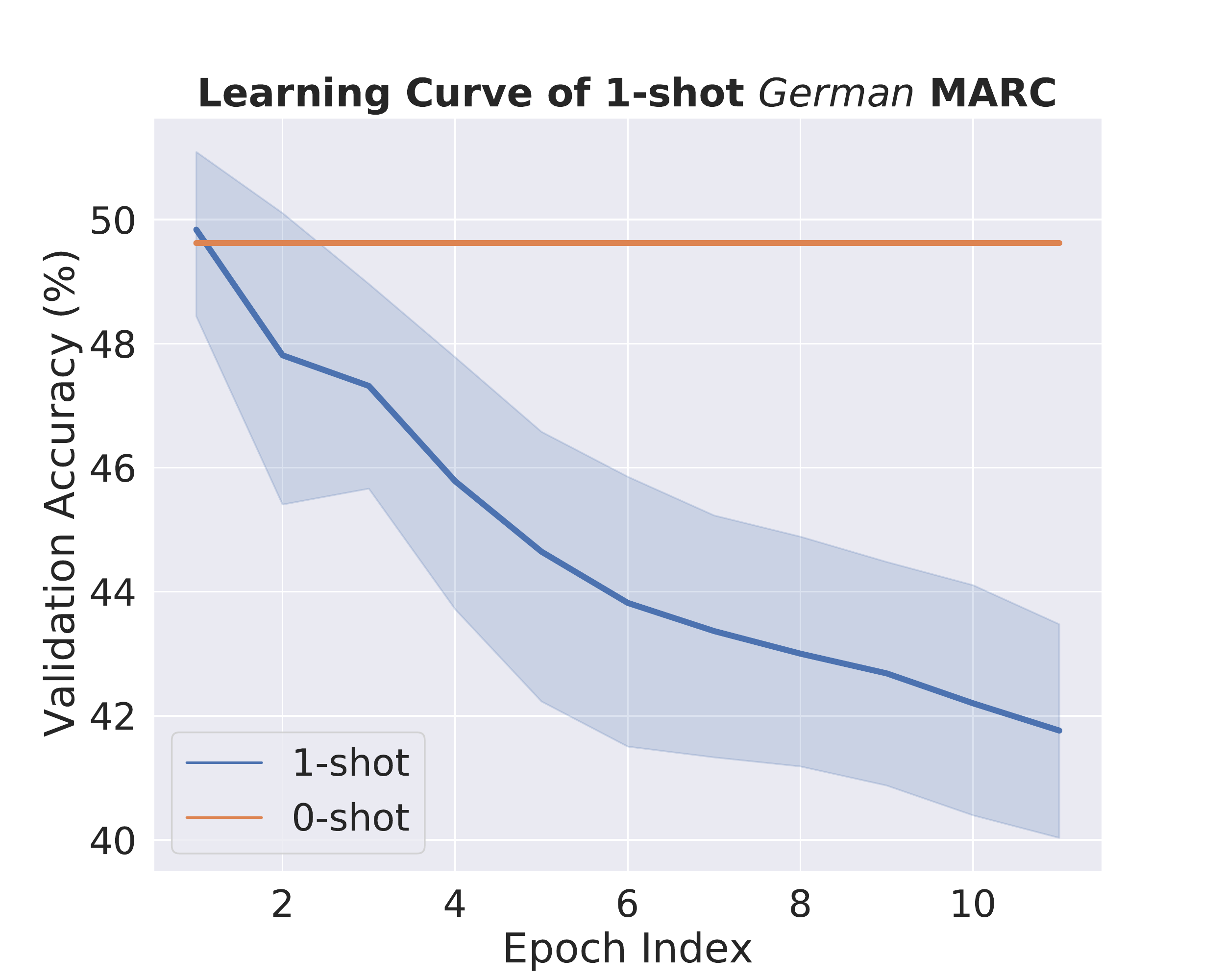}
\caption{
     Early stopped 1-shot transfer (EN $\to$ DE) learning curve.
     The English-trained model overfits the 1-shot bucket quickly,
     showing decreasing dev performance during training.   
  }
\figlabel{delearningcurve}
\end{figure}

\input{lexicaloverlap}

%% file: dataexampletableappendix.tex
\begin{table*}[t]
\scriptsize
\centering
\renewcommand{\arraystretch}{1.1}
\setlength{\tabcolsep}{4pt}   
\begin{tabular}{lll}
\hline
\multirow{2}{*}{MARC}  & Text A & Très mignons et de bonne qualité. La figurine est assez imposante mais conforme à la taille indiquée dans le descriptif. \\
                       & Text B & Jolis détails . home                     \\ \hline
\multirow{2}{*}{XNLI}  & Text A & Ich musste anfagen Seminare zu belegen . \\
                       & Text B & Ich brauchte keine Vorbereitung .        \\ \hline
\multirow{2}{*}{PAWSX} & Text A & Lo entrenó John Velázquez y en sus carreras más importantes lo montó el jinete Dale Romans.                             \\
                       & Text B & Lo entrenó John Velázquez, y el jinete Dale Romans lo montó en las carreras más importantes.                            \\\hline
\multirow{1}{*}{POS}   & Text A & (Lo,PRON), (sanno,VERB), (oramai,ADV), (quasi,ADV), (tutti,PRON), (che,SCONJ), (un,DET), (respiro,NOUN), (affannoso,ADJ) ...       \\ \hline
\multirow{1}{*}{NER}   & Text A & (Sempat,O), (pindah,O), (ke,O), (HJK,B-ORG), (dan,O), (1899,B-ORG), (Hoffenheim,I-ORG), (yang,O), (meminjamkannya,O), (ke,O) ... \\ \hline
\end{tabular}
\caption{Example entries of the datasets.
  We convert the raw text to the mBERT format ``Text A'' and ``Text B'' \citep{devlin-etal-2019-bert}.
  For POS and NER, we list (word, tag) pairs in the sentence.
  Following \citet{SCHWENK18.658}, we provide document indices of MLDoc
  for retrieving the documents from RCV1 and RCV2.}
\tablabel{datasetexamples}
\end{table*}

%% file: linguistictable.tex
\begin{table*}[t]
\centering
\scriptsize
\renewcommand{\arraystretch}{1.2}
\setlength{\tabcolsep}{1.4pt}
\resizebox{!}{5.2cm}{
\begin{tabular}{c|c|cccccc}
\multirow{2}{*}{Language} & \multirow{2}{*}{Writing Script} & 81A & 85A & 86A & 87A & 88A & 89A\\
                          && Order of Subject, Object and Verb
                          & Order of adposition and noun 
                          & Order of genitive and noun 
                          & Order of adjective and noun
                          & Order of demonstrative and noun 
                          & Order of numeral and noun\\ \hline
English (\textsc{en}) & Latin & SVO & Prepositions & No dominant order & Adjective-noun & Demonstrative-noun & Numeral-noun\\ 
Afrikaans (\textsc{af}) & Latin & - & - & - & - & - & -\\
Arabic (\textsc{ar}) & Arabic & VSO & Prepositions & Noun-genetive & Noun-adjective & Demonstrative-noun & Numeral-noun\\
Bulgarian (\textsc{bg}) & Cyrillic & SVO & Prepositions & No dominant order & Adjective-noun & Demonstrative-noun & Numeral-noun\\
Bengali (\textsc{bn}) & Brahmic & SOV & - & - & - & - & -\\
German (\textsc{de}) & Latin & No dominant order & Prepositions & Noun-genetive & Adjective-noun & Demonstrative-noun & Numeral-noun\\
Greek (\textsc{el}) & Greek & No dominant order & Prepositions & Noun-genetive & Adjective-noun & Demonstrative-noun & Numeral-noun\\
Spanish (\textsc{es}) & Latin & SVO & Prepositions & Noun-genetive & Noun-adjective & Demonstrative-noun & Numeral-noun\\
Estonian (\textsc{et}) & Latin & SVO & Postpositions & Genetive-noun & Adjective-noun & Demonstrative-noun & Numeral-noun\\
Basque (\textsc{eu}) & Latin & SOV & Postpositions & Genetive-noun & Noun-adjective & Noun-demonstrative & Numeral-noun\\
Persian (\textsc{fa}) & Perso-Arabic & SOV & Prepositions & Noun-genetive & Noun-adjective & Demonstrative-noun & Numeral-noun\\
Finnish (\textsc{fi}) & Latin & SVO & Postpositions & Genetive-noun & Adjective-noun & Demonstrative-noun & Numeral-noun\\
French (\textsc{fr}) & Latin & SVO & Prepositions & Noun-genetive & Noun-adjective & Demonstrative-noun & Numeral-noun\\
Hebrew (\textsc{he}) & Hebrew & SVO & Prepositions & Noun-genetive & Noun-adjective & Noun-demonstrative & Numeral-noun\\
Hindi (\textsc{hi}) & Devanagari & SOV & Postpositions & Genetive-noun & Adjective-noun & Demonstrative-noun & Numeral-noun\\
Hungarian (\textsc{hu}) & Latin & No dominant order & Postpositions & Genetive-noun & Adjective-noun & Demonstrative-noun & Numeral-noun\\
Indonesian (\textsc{id}) & Latin & SVO & Prepositions & Noun-genetive & Noun-adjective & Noun-demonstrative & Numeral-noun\\
Italian (\textsc{it}) & Latin & SVO & Prepositions & Noun-genetive & Noun-adjective & Demonstrative-noun & Numeral-noun\\
Japanese (\textsc{ja}) & Ideograms  & SOV & Postpositions & Genetive-noun & Adjective-noun & Demonstrative-noun & Numeral-noun\\
Javanese (\textsc{jv}) & Latin & - & - & - & - & - & -\\
Georgian (\textsc{ka}) & Georgian & SOV & Postpositions & Genetive-noun & Adjective-noun & Demonstrative-noun & Numeral-noun\\
Kazakh (\textsc{kk}) & Cyrillic & - & - & - & - & - & -\\
Korean (\textsc{ko}) & Hangul & SOV & Postpositions & Genetive-noun & Adjective-noun & Demonstrative-noun & Numeral-noun\\
Malayalam (\textsc{ml}) & Brahmic & SOV & - & Genetive-noun & Adjective-noun & Demonstrative-noun & Numeral-noun\\
Marathi (\textsc{mr}) & Devanagari & SOV & Postpositions & Genetive-noun & Adjective-noun & Demonstrative-noun & Numeral-noun\\
Malay (\textsc{ms}) & Latin & - & - & - & - & - & -\\
Burmese (\textsc{my}) & Brahmic & SOV & Postpositions & Genetive-noun & Noun-adjective & Demonstrative-noun & Noun-numeral\\
Dutch (\textsc{nl}) & Latin & No dominant order & Prepositions & Noun-genetive & Adjective-noun & Demonstrative-noun & Numeral-noun\\
Portuguese (\textsc{pt}) & Latin & SVO & Prepositions & Noun-genetive & Noun-adjective & Demonstrative-noun & -\\
Russian (\textsc{ru}) & Cyrillic & SVO & Prepositions & Noun-genetive & Adjective-noun & Demonstrative-noun & Numeral-noun\\
Swahili (\textsc{sw}) & Latin & SVO & Prepositions & Noun-genetive & Noun-adjective & Noun-demonstrative & Noun-numeral\\
Tamil (\textsc{ta}) & Brahmic & SOV & Postpositions & Genetive-noun & Adjective-noun & Demonstrative-noun & Numeral-noun\\
Telugu (\textsc{te}) & Brahmic & SOV & Postpositions & Genetive-noun & Adjective-noun & Demonstrative-noun & Numeral-noun\\
Thai (\textsc{th}) & Brahmic & SVO & Prepositions & Noun-genetive & Noun-adjective & Noun-demonstrative & Noun-numeral\\
Tagalog (\textsc{tl}) & Latin & VSO & - & Noun-genetive & No dominant order & Mixed & Numeral-noun\\
Turkish (\textsc{tr}) & Latin & SOV & Postpositions & Genetive-noun & Adjective-noun & Demonstrative-noun & Numeral-noun\\
Urdu (\textsc{ur}) & Perso-Arabic & SOV & Postpositions & Genetive-noun & Adjective-noun & Demonstrative-noun & Numeral-noun\\
Vietnamese (\textsc{vi}) & Latin & SVO & Prepositions & Noun-genetive & Noun-adjective & Noun-demonstrative & Numeral-noun\\
Yoruba (\textsc{yo}) & Latin & SVO & Prepositions & Noun-genetive & Noun-adjective & Noun-demonstrative & Noun-numeral\\ 
Chinese (\textsc{zh}) & Chinese ideograms & SVO & No dominant order & Genetive-noun & Adjective-noun & Demonstrative-noun & Numeral-noun\\ \hline
\end{tabular}}
\caption{All languages for the experiments along with their ISO 639-1 codes, writing script, and linguistic features. ``-" denotes lacking feature information from WALS.}
\tablabel{linguisticfeatures}
\end{table*}

%% file: minincl.tex
\begin{algorithm}[t]
	\caption{Minimum-including}\label{miniincludealgorithm}
	\footnotesize
	\begin{algorithmic}[1]
		\Require \# of shot $K$, language data $\mathcal{D}$, label set $\mathcal{L_D}$ \\
		Initialize a bucket $\mathcal{S}=\left\{ \right\}$, $\text{Count}_{\ell_j} = 0 $ 
		$(\forall \ell_j \in \mathcal{L_D})$
		\\
		\For{$\ell$ in $\mathcal{L_D}$} {
			\While{$\text{Count}_{\ell} < K $ }
			{From $\mathcal{D}$, randomly sample a $(\bm{x}^{(i)},\bm{y}^{(i)})$ pair that $\bm{y}^{(i)}$ includes $\ell$
				
				Add $(\bm{x}^{(i)},\bm{y}^{(i)})$ to $\mathcal{S}$
				
				Update all $\text{Count}_{\ell_j}$ 
				$(\forall \ell_j \in \mathcal{L_D})$
			}
		} \\
		\For{each $(\bm{x}^{(i)},\bm{y}^{(i)})$ in $\mathcal{S}$}
		{   
			Remove $(\bm{x}^{(i)},\bm{y}^{(i)})$ from $\mathcal{S}$ 
			
			Update all $\text{Count}_{\ell_j}$ 
			$(\forall \ell_j \in \mathcal{L_D})$ 
			
			\If{any $\text{Count}_{\ell_j} < $  K}
			{Put $(\bm{x}^{(i)},\bm{y}^{(i)})$ back to $\mathcal{S}$
				
				Update all $\text{Count}_{\ell_j}$ 
				$(\forall \ell_j \in \mathcal{L_D})$
			}
		}\\
		Return $\mathcal{S}$ 
	\end{algorithmic}
      \end{algorithm}

%% file: posfewshottable.tex
\begin{table}[!]
  \centering
  \tiny
    \def\arraystretch{0.93}  
  \begin{tabular}{ccccc}
                 & K=0   & K=1               & K=2              & K=4                                 \\\hline
    EN  & 95.39 & -                 & -                & -                                   \\
    AF  & 86.60 & 91.10 $\pm$ 1.11 & 92.12 $\pm$ 1.15 & 93.50 $\pm$ 0.56 \\
    AR  & 66.55 & 75.64 $\pm$ 1.09 & 77.01 $\pm$ 0.84 & 78.52 $\pm$ 0.67 \\
    BG  & 87.02 & 91.01 $\pm$ 0.97 & 91.97 $\pm$ 0.90 & 93.18 $\pm$ 0.56 \\
    DE  & 86.38 & 89.38 $\pm$ 0.90 & 90.21 $\pm$ 0.50 & 91.32 $\pm$ 0.43 \\
    EL  & 81.89 & 89.69 $\pm$ 1.05 & 90.53 $\pm$ 0.89 & 91.58 $\pm$ 0.72 \\
    ES  & 86.64 & 90.05 $\pm$ 1.01 & 91.19 $\pm$ 0.74 & 92.31 $\pm$ 0.52 \\
    ET  & 79.17 & 81.69 $\pm$ 1.09 & 83.05 $\pm$ 0.98 & 84.39 $\pm$ 0.56 \\
    EU  & 49.51 & 68.44 $\pm$ 2.47 & 71.94 $\pm$ 1.78 & 75.89 $\pm$ 1.20 \\
    FA  & 65.73 & 80.82 $\pm$ 2.14 & 82.81 $\pm$ 1.79 & 84.95 $\pm$ 1.16 \\
    FI  & 74.49 & 78.25 $\pm$ 1.22 & 79.65 $\pm$ 0.85 & 81.32 $\pm$ 0.82 \\
    FR  & 82.54 & 89.55 $\pm$ 1.08 & 90.84 $\pm$ 0.64 & 91.66 $\pm$ 0.60 \\
    HE  & 76.79 & 80.40 $\pm$ 1.42 & 82.42 $\pm$ 1.06 & 83.98 $\pm$ 0.83 \\
    HI  & 64.29 & 78.87 $\pm$ 1.26 & 80.80 $\pm$ 0.80 & 81.97 $\pm$ 0.92 \\
    HU  & 75.10 & 84.44 $\pm$ 1.40 & 86.31 $\pm$ 0.90 & 88.61 $\pm$ 0.67 \\
    ID  & 70.80 & 72.68 $\pm$ 1.08 & 73.64 $\pm$ 0.78 & 74.34 $\pm$ 0.75\\
    IT  & 85.97 & 88.77 $\pm$ 0.87 & 89.93 $\pm$ 0.50 & 90.77 $\pm$ 0.59\\
    JA  & 47.60 & 75.84 $\pm$ 1.68 & 78.46 $\pm$ 1.31 & 80.42 $\pm$ 0.98 \\
    KO  & 42.29 & 57.43 $\pm$ 1.36 & 59.92 $\pm$ 1.18 & 62.37 $\pm$ 1.22 \\
    MR  & 58.70 & 71.60 $\pm$ 2.52 & 74.89 $\pm$ 1.95 & 77.21 $\pm$ 1.77 \\
    NL  & 88.35 & 88.97 $\pm$ 0.73 & 89.55 $\pm$ 0.79 & 90.83 $\pm$ 0.54 \\
    PT  & 86.45 & 88.18 $\pm$ 0.70 & 88.98 $\pm$ 0.66 & 89.78 $\pm$ 0.38\\
    RU  & 86.36 & 89.07 $\pm$ 0.76 & 89.85 $\pm$ 0.57 & 91.13 $\pm$ 0.51  \\
    TA  & 53.51 & 62.84 $\pm$ 2.69 & 66.30 $\pm$ 1.56 & 69.36 $\pm$ 1.13 \\
    TE  & 67.48 & 71.46 $\pm$ 2.58 & 75.72 $\pm$ 1.94 & 78.84 $\pm$ 1.44 \\
    TR  & 57.58 & 64.01 $\pm$ 1.53 & 66.02 $\pm$ 1.28 & 67.73 $\pm$ 0.82 \\
    UR  & 52.40 & 74.95 $\pm$ 2.15 & 78.53 $\pm$ 1.38 & 79.57 $\pm$ 1.24 \\
    VI  & 54.96 & 64.79 $\pm$ 2.33 & 69.39 $\pm$ 1.73 & 72.36 $\pm$ 1.51 \\
    ZH  & 63.01 & 74.15 $\pm$ 1.96 & 76.62 $\pm$ 1.39 & 79.42 $\pm$ 0.83
  \end{tabular}
   \caption{Zero- (column K$=$0) and few- (columns K$>$0)
    shot cross-lingual transfer results (\%) on POS test set.}
  \tablabel{posfewshot}
\end{table}

%% file: nerfewshottable.tex
\begin{table}[!]
  \centering
  \tiny
    \def\arraystretch{0.93}  
  \begin{tabular}{ccccc}
                & K=0         & K=1               & K=2             & K=4              \\ \hline
    EN & 83.65       & -                 & -               & -                \\
    AF & 78.36       & 79.07 $\pm$ 1.47  &79.69 $\pm$ 1.40 &80.24 $\pm$ 1.16  \\
    AR & 39.91       & 54.44 $\pm$ 6.74  &60.51 $\pm$ 4.30 &63.61 $\pm$ 2.65  \\
    BG & 78.59       & 78.65 $\pm$ 0.38  &78.70 $\pm$ 0.39 &78.87 $\pm$ 0.48  \\
    BN & 64.17       & 66.37 $\pm$ 1.69  &66.66 $\pm$ 1.57 &65.98 $\pm$ 2.11  \\
    DE & 79.00       & 79.33 $\pm$ 0.71  &79.61 $\pm$ 0.76 &79.74 $\pm$ 0.73  \\
    EL & 75.20       & 74.93 $\pm$ 0.79  &75.18 $\pm$ 0.95 &75.40 $\pm$ 0.93  \\
    ES & 77.16       & 79.19 $\pm$ 1.97  &80.28 $\pm$ 1.71 &80.90 $\pm$ 1.94  \\
    ET & 71.88       & 72.58 $\pm$ 1.17  &73.60 $\pm$ 1.65 &74.60 $\pm$ 1.59  \\
    EU & 55.35       & 59.60 $\pm$ 3.32  &61.59 $\pm$ 3.84 &64.68 $\pm$ 2.96  \\
    FA & 40.73       & 59.20 $\pm$ 5.34  &68.55 $\pm$ 4.04 &71.13 $\pm$ 3.45  \\
    FI & 68.43       & 71.43 $\pm$ 2.61  &73.92 $\pm$ 2.44 &75.81 $\pm$ 2.15  \\
    FR & 80.38       & 80.54 $\pm$ 0.93  &81.08 $\pm$ 0.85 &81.22 $\pm$ 0.93  \\
    HE & 56.36       & 58.24 $\pm$ 2.25  &59.43 $\pm$ 2.29 &60.27 $\pm$ 2.43  \\
    HI & 65.84       & 67.16 $\pm$ 1.61  &67.56 $\pm$ 2.18 &68.29 $\pm$ 1.76  \\
    HU & 71.28       & 72.23 $\pm$ 1.33  &73.03 $\pm$ 1.44 &74.14 $\pm$ 1.61  \\
    ID & 60.10       & 77.87 $\pm$ 6.31  &78.57 $\pm$ 4.14 &81.07 $\pm$ 1.50  \\
    IT & 80.30       & 80.68 $\pm$ 0.79  &81.00 $\pm$ 0.92 &80.90 $\pm$ 1.12  \\
    JA & 7.16        & 20.71 $\pm$ 7.07  &28.23 $\pm$ 5.32 &32.93 $\pm$ 6.03  \\
    JV & 61.18       & 67.80 $\pm$ 4.72  &69.79 $\pm$ 3.37 &72.12 $\pm$ 3.34  \\
    KA & 61.26       & 61.62 $\pm$ 1.09  &62.25 $\pm$ 1.56 &63.68 $\pm$ 1.66  \\
    KK & 40.29       & 50.42 $\pm$ 5.49  &54.97 $\pm$ 6.81 &62.94 $\pm$ 4.55  \\
    KO & 46.50       & 47.25 $\pm$ 1.36  &48.69 $\pm$ 1.82 &51.76 $\pm$ 2.30  \\
    ML & 46.77       & 47.83 $\pm$ 2.30  &49.51 $\pm$ 3.01 &51.41 $\pm$ 3.31  \\
    MR & 54.70       & 55.78 $\pm$ 2.54  &57.22 $\pm$ 2.43 &59.18 $\pm$ 3.13  \\
    MS & 68.61       & 71.04 $\pm$ 3.07  &74.51 $\pm$ 4.28 &76.25 $\pm$ 3.04  \\
    MY & 42.45       & 43.55 $\pm$ 3.88  &46.03 $\pm$ 4.48 &47.81 $\pm$ 4.28  \\
    NL & 82.77       & 82.73 $\pm$ 0.43  &82.83 $\pm$ 0.54 &82.82 $\pm$ 0.46  \\
    PT & 79.28       & 79.89 $\pm$ 0.99  &80.39 $\pm$ 0.98 &80.49 $\pm$ 0.95  \\
    RU & 65.20       & 67.30 $\pm$ 2.38  &68.78 $\pm$ 2.73 &71.34 $\pm$ 2.82  \\
    SW & 68.36       & 71.07 $\pm$ 4.28  &70.08 $\pm$ 3.15 &74.33 $\pm$ 5.25  \\
    TA & 46.12       & 47.81 $\pm$ 1.81  &49.86 $\pm$ 2.99 &52.23 $\pm$ 2.63  \\
    TE & 50.02       & 52.57 $\pm$ 1.91  &54.02 $\pm$ 2.65 &55.75 $\pm$ 2.72  \\
    TH & 1.53        & 4.56 $\pm$ 4.87   &6.08 $\pm$ 4.88  &5.87 $\pm$ 4.14   \\
    TL & 69.23       & 72.34 $\pm$ 2.25  &72.63 $\pm$ 2.43 &73.55 $\pm$ 2.25  \\
    TR & 65.78       & 69.37 $\pm$ 2.24  &69.53 $\pm$ 2.07 &72.33 $\pm$ 2.85  \\
    UR & 40.77       & 58.48 $\pm$ 6.51  &63.38 $\pm$ 4.88 &66.49 $\pm$ 4.64  \\
    VI & 64.67       & 68.77 $\pm$ 3.54  &69.64 $\pm$ 3.63 &71.08 $\pm$ 3.28  \\
    YO & 35.48       & 53.55 $\pm$ 6.19  &58.22 $\pm$ 5.47 &65.46 $\pm$ 7.10  \\
    ZH & 13.95       & 32.84 $\pm$ 7.10  &40.34 $\pm$ 5.32 &48.49 $\pm$ 4.30
  \end{tabular}
  \caption{Zero- (column K$=$0) and few- (columns K$>$0)
    shot cross-lingual transfer
    results (\%) on NER test set.}
\tablabel{nerfewshot}
\end{table}

%% file: lexicaloverlap.tex
\begin{table}[h!]
  \centering
  \scriptsize
    \def\arraystretch{0.93}  
  \begin{tabular}{ccccccc}
         & \multicolumn{3}{c}{NER} & \multicolumn{3}{c}{POS}\\\cmidrule(lr){2-4}\cmidrule(lr){5-7}
         & K=1               & K=2              & K=4  & K=1               & K=2              & K=4                                 \\\hline
        AF & 4.54 & 8.75 & 13.44 & 4.97 & 6.11 & 7.90 \\
        AR & 0.65 & 0.95 & 1.57 & 3.51 & 4.49 & 5.30 \\
        BG & 0.98 & 2.19 & 3.23 &-&-&-\\
        BN & 0.39 & 0.77 & 0.80 &-&-&-\\
        DE & 8.75 & 13.20 & 20.61 & 9.36 & 15.33 & 21.48 \\
        EL & 1.45 & 1.84 & 3.59 & 1.96 & 2.87 & 3.04 \\
        ES & 6.29 & 10.59 & 19.66 & 10.00 & 17.53 & 22.63 \\
        ET & 4.80 & 5.96 & 11.24 & 5.81 & 9.22 & 13.17 \\
        EU & 3.77 & 5.55 & 12.31 & 2.60 & 3.45 & 4.69 \\
        FA & 0.27 & 0.44 & 1.01 & 0.37 & 0.37 & 0.41 \\
        FI & 5.61 & 9.05 & 15.66 & 4.59 & 7.03 & 8.78 \\
        FR & 6.26 & 10.83 & 19.01 & 15.60 & 25.23 & 37.39 \\
        HE & 0.86 & 1.90 & 3.23 & 1.22 & 1.93 & 2.26 \\
        HI & 0.95 & 1.16 & 1.99 & 0.44 & 0.27 & 0.51 \\
        HU & 5.07 & 9.19 & 14.35 & 3.18 & 3.92 & 4.15 \\
        ID & 5.34 & 9.82 & 16.94 & 9.39 & 13.78 & 21.75 \\
        IT & 7.89 & 10.94 & 21.27 & 11.99 & 16.15 & 21.35 \\
        JA & 1.75 & 2.02 & 2.14 & 2.60 & 3.68 & 5.00 \\
        JV & 2.49 & 3.05 & 3.44 &-&-&-\\
        KA & 1.99 & 4.00 & 5.78 &-&-&-\\
        KK & 0.89 & 1.22 & 2.11 &-&-&-\\
        KO & 1.48 & 1.54 & 3.32 & 2.33 & 3.85 & 5.67 \\
        ML & 0.36 & 1.04 & 1.30 &-&-&-\\
        MR & 0.53 & 0.56 & 0.71 & 0.24 & 0.24 & 0.24 \\
        MS & 4.86 & 7.44 & 13.70 &-&-&-\\
        MY & 0.21 & 0.36 & 0.42 &-&-&-\\
        NL & 7.18 & 10.65 & 20.14 & 7.94 & 11.42 & 16.79 \\
        PT & 6.29 & 11.00 & 19.13 & 8.88 & 13.38 & 20.13\\
        RU & 1.60 & 2.34 & 3.77 & 4.15 & 6.11 & 9.32 \\
        SW & 5.90 & 8.10 & 12.37 &-&-&-\\
        TA & 0.65 & 1.54 & 2.08 & 1.32 & 1.28 & 1.62 \\
        TE & 0.77 & 0.80 & 1.19 & 0.20 & 0.20 & 0.20 \\
        TH & 1.63 & 1.87 & 2.08 &-&-&- \\
        TL & 4.83 & 8.96 & 14.98 &-&-&- \\
        TR & 4.89 & 8.48 & 16.43 & 2.09 & 2.26 & 3.01 \\
        UR & 0.30 & 0.27 & 0.68 & 0.74 & 1.35 & 2.16 \\
        VI & 4.33 & 8.39 & 13.41 & 1.62 & 2.16 & 2.90 \\
        YO & 1.90 & 2.58 & 2.88 &-&-&- \\
        ZH & 1.81 & 1.99 & 2.14 & 3.04 & 4.86 & 7.33 \\
  \end{tabular}
  \caption{Lexical overlap (per-mille) of target languages with EN for NER and POS using different K-shot buckets.}
  \tablabel{lexicaloverlap}
\end{table}